\documentclass{article} %
\usepackage{iclr2025_conference,times}

\usepackage{amsmath,amsfonts,bm}

\def\eqref#1{equation~\ref{#1}}

\def\1{\bm{1}}

\DeclareMathAlphabet{\mathsfit}{\encodingdefault}{\sfdefault}{m}{sl}
\SetMathAlphabet{\mathsfit}{bold}{\encodingdefault}{\sfdefault}{bx}{n}

\usepackage{hyperref}
\usepackage{url}
\usepackage{booktabs}
\usepackage{graphicx}
\usepackage{tabularx}
\usepackage{multirow}
\usepackage{colortbl}
\usepackage{bbm}
\usepackage{amssymb}
\usepackage{siunitx}
\newtheorem{problem}{Problem}

\newcommand{\angstrom}{\textup{\r{A}}}

\title{FreeCG: Free the Design Space of Clebsch–Gordan Transform for Machine Learning Force Fields}

\author{Shihao Shao$^1$\thanks{Please address correspondence to us at shaoshihao@pku.edu.cn or cuiqinghua@hsc.pku.edu.cn} \; Haoran Geng$^1$ \; Zun Wang$^2$ \; Qinghua Cui$^1$$^*$\\
$^1$Peking University \; $^2$Microsoft \\
}

\iclrfinalcopy %
\begin{document}

\maketitle
\vspace{-0.5cm}
\begin{abstract}
\vspace{-0.3cm}
Machine Learning Force Fields (MLFFs) are of great importance for chemistry, physics, materials science, and many other related fields. The Clebsch–Gordan Transform (CG transform) effectively encodes many-body interactions and is thus an important building block for many models of MLFFs. However, the permutation-equivariance requirement of MLFFs limits the design space of CG transform, that is, intensive CG transform has to be conducted for each neighboring edge and the operations should be performed in the same manner for all edges. This constraint results in reduced expressiveness of the model while simultaneously increasing computational demands.
To overcome this challenge, we first implement the CG transform layer on the permutation-invariant abstract edges generated from real edge information. We show that this approach allows complete freedom in the design of the layer without compromising the crucial symmetry. Developing on this free design space, we further propose group CG transform with sparse path, abstract edges shuffling, and attention enhancer to form a powerful and efficient CG transform layer. Our method, known as \textit{\textbf{FreeCG}}, achieves state-of-the-art (SOTA) results in force prediction for MD17, rMD17, MD22, and is well extended to property prediction in QM9 datasets with several improvements greater than 15$\%$ and the maximum beyond 20$\%$. The extensive real-world applications showcase high practicality. FreeCG introduces a novel paradigm for carrying out efficient and expressive CG transform in future geometric neural network designs. To demonstrate this, the recent SOTA, QuinNet, is also enhanced under our paradigm. Code will be publicly available.
\end{abstract}

\vspace{-0.55cm}
\section{Introduction}\label{sec:intro}
\vspace{-0.3cm}

Machine Learning Force Fields (MLFFs) are of great importance for drug development~\citep{chen2024design}, materials science~\citep{liu2024layer}, chemical reaction kinetics \citep{meuwly2021machine}, nanotechnology \citep{wang2023novel}, among others. It offers a satisfactory trade-off between accuracy and efficiency, which is expected to perform as powerful as Density Functional Theory (DFT) \citep{kohn1965self} or other high accuracy references \citep{martin2020electronic,ceperley1980ground,bartlett2007coupled}, but with orders-of-magnitude speedup \citep{cui2024geometry,wang2023quinnet,wang2024enhancing,musaelian2023learning,batzner20223,drautz2019atomic,batatia2022mace,tholke2021equivariant,schutt2018schnet,chmiela2017machine}.

Graph Neural Networks (GNNs) perform SOTA on several MLFFs benchmarks \citep{schutt2018schnet,schutt2021equivariant}. Group and group representation theory play important roles in the design of GNNs for MLFFs. Rotation invariance is generally required in these works, as we naturally require the potential energy unchanged w.r.t. rotations of the input molecule. A recent design trend is to maintain rotation, reflection, and translation equivariance in the design of neural networks. They hope the internal features can move with respect to the input molecule, enabling higher expressive power. GNNs that obey this property are called Equivariant Graph Neural Networks (EGNNs) \citep{thomas2018tensor,satorras2021n,gasteiger2020directional,vaswani2017attention,fuchs2020se,liao2022equiformer}. 
To better model many-body interactions, irreducible representations (irreps) are adopted to represent high-order geometric objects. In this context, the Clebsch-Gordan (CG) transform is used to translate between different irreps.
Several works leverage such high degree irreps or tensors, showing significant performance boost \citep{batatia2022mace,batzner20223,musaelian2023learning,gasteiger2021gemnet,thomas2018tensor,simeon2024tensornet}. However, the benefit of high degree irreps and CG transform performed on them is at the cost of heavy computational overhead. The reason is, tensors are extensions of scalars and vectors, and in this way CG transform also extends the dot product. Thus, the higher the degree of irreps for the CG transform, the greater the computational demands. The requirements for being permutation equivariant make this burden hard to alleviate. Unlike rotation or translation equivariance, permutation equivariance is often implicitly guaranteed in EGNNs, which means the order of internal features should changes according to it of input atoms. To maintain permutation equivariance, EGNNs require each node to receive information from neighboring atoms together with the edges linking them, where the heavy computation of CG transform occurs for each neighboring atom and edge. This means we cannot naïvely remove some neighbor computations, as it will break permutation equivariance. Moreover, the narrowness of the design space prevents us from freely constructing the CG transform layer, and thus limits the expressivity of models. For instance, we need to operate on each neighboring atom in an equal way (\textit{e.g.,} the predecessors typically assign a same MLP operating on scalar features of the edge to produce the weights for each computation between the central atom and each neighboring one \citep{batzner20223,musaelian2023learning}).

In this work, to confront this challenge, we propose \textit{\textbf{FreeCG}}. The model generates and refines geometric features from the surrounding edges near each atom. We call the different aggregated edge geometric features \textit{abstract edges}, which are permutation invariant when we consider the internal features maintained by a given atom w.r.t. the neighbouring atoms and edges. By the invariance transitivity, we show that CG transform on these abstract edges is also permutation invariant, regardless of designs, and does not affect the permutation equivariance of the layer, thus being free of the burdens above. Furthermore, the abstract edges are constructed from different real edges, so they contain refined features of them for better model expressive power. The invariance nature of abstract edges allows us to assign different weights to different edges, instead of weights computed by the same MLP. The free design space allows us to do more. We put abstract edges into groups, and operate on each group individually, to further decrease the computation demands. Previous works that keep E(3)-equivariance are more expensive \citep{batzner20223,musaelian2023learning}, as they require an extra \textit{parity} argument being $1$ or $-1$, and thus the number of irreps is doubled. Instead, we select an efficient set of paths for CG transform so that we maintain E(3)-equivariance while being more efficient than keeping SE(3)-equivariance. The abstract edges shuffling, inspired by \citep{zhang2018shufflenet}, is also implied for combination of irreps features. The abstract edges are then plugged back into the cross-attention calculation to improve the quality of the attention scores. The operations above are available thanks to the invariance properties of the abstract edges. 

\vspace{-0.1cm}

The contributions are summarized as follows:

\vspace{-0.1cm}

\textbf{1)} We propose the concept of \textbf{abstract edges}, by which we design several new modules to resolve a major challenge in the current EGNNs: the narrowness design space of CG transform, which results in reduced expressivity and high computation overhead.

\vspace{-0.1cm}

%to leverage the permutation-invariant transitivity and abstract edges 
% to completely free the design space of CG transform. 
\textbf{2)} We propose \textit{\textbf{FreeCG}}, comprising of three main components: \textit{\textbf{Group CG transform with sparse path}}, \textit{\textbf{abstract edges shuffling}}, and \textit{\textbf{Attention enhancer}}. These contribute to an informative and efficient model with high-order irreps and CG transform. 

\vspace{-0.1cm}

\textbf{3)} Experiments on small molecule datasets MD17, rMD17, large molecules ones MD22, and molecular property datasets QM9 reveal the SOTA performance of FreeCG, with several improvements beyond 15\%. The real-world applications further indicate the practicality. 

\vspace{-0.1cm}

\textbf{4)} This work presents a new paradigm for CG transform in future research, extending beyond the design presented here. We further enhance QuinNet under this paradigm to demonstrate this point.

\vspace{-0.5cm}
\section{Related Works}
\vspace{-0.4cm}

Maintaining E(3)-/SE(3)-equivariance has become a popular trend in the design of neural networks for MLFFs. The challenge lies in how to properly construct geometric objects that are better capable of modeling the atomic environment. In general, the methods can be categorized into the following two lines:

\noindent\textbf{Methods with geometric vectors.} Several methods directly work on regular geometric vectors \citep{schutt2021equivariant,schutt2018schnet,gasteiger2020directional,gasteiger2020dimenetpp}. The starting point of this line of works is about covering necessary information for MLFFs. For instance, SchNet \citep{schutt2018schnet} first introduces continuous-filter to MLFFs, but only distance information is covered. DimeNet \citep{gasteiger2020directional} further considers bond angle information to obtain higher capacity.
To make models capable of capturing many-body interactions, most of the following models in this line explicitly encode such types of interactions \citep{tholke2021equivariant,wang2024enhancing,wang2023quinnet}. This is commonly done via calculating different angles between atoms or surfaces (e.g., torsion angle, improper angle \citep{wang2024enhancing}, dihedral angle \citep{wang2023quinnet}).
The irreducible representations (irreps) and Clebsch-Gordan (CG) transform, on the other hand, can implicitly encode many-body interactions in well-defined mathematical objects. This work presents FreeCG with highly efficient and expressive CG transform layers, significantly outperforming previous works by maximum margins and with minor computational load.

\noindent\textbf{Methods equipped with irreps and CG transform.} Applying irreps and CG transform to Equivariant Graph Neural Networks (EGNNs) was first proposed as more of a conceptual framework, known as Tensor Field Network \citep{thomas2018tensor}. Several works have been proposed on top of this foundation. For instance, NequIP \citep{batzner20223} has implemented the idea to construct high-order irreps, and shows state-of-the-art (SOTA) results for MLFFs. Allegro \citep{musaelian2023learning} resolves the challenges of the scaling issue, which makes it possible to run parallelly on a large number of GPUs. SE(3)-Transformer \citep{fuchs2020se} first proposes equivariant dot-product for generating self-attention. Equiformer \citep{liao2022equiformer} further achieves E(3)-equivariance combining MLP attention and non-linear messages. MACE \citep{batatia2022mace} extends classic body order expansion methods, Atomic Cluster Expansion (ACE) \citep{drautz2019atomic}, to a hierarchical framework. 
CG transform is a fundamental building block in these works, but with a limited design space, affecting both performance and efficiency. In this work, we completely free the design space of CG transform and propose a novel model, FreeCG, performing strong SOTA and showing high efficiency for MLFFs, which is also a new design paradigm for future works.

\begin{figure*}[tb]
  \label{fig:teaset}
  \centering
  \includegraphics[width=1.0\linewidth]{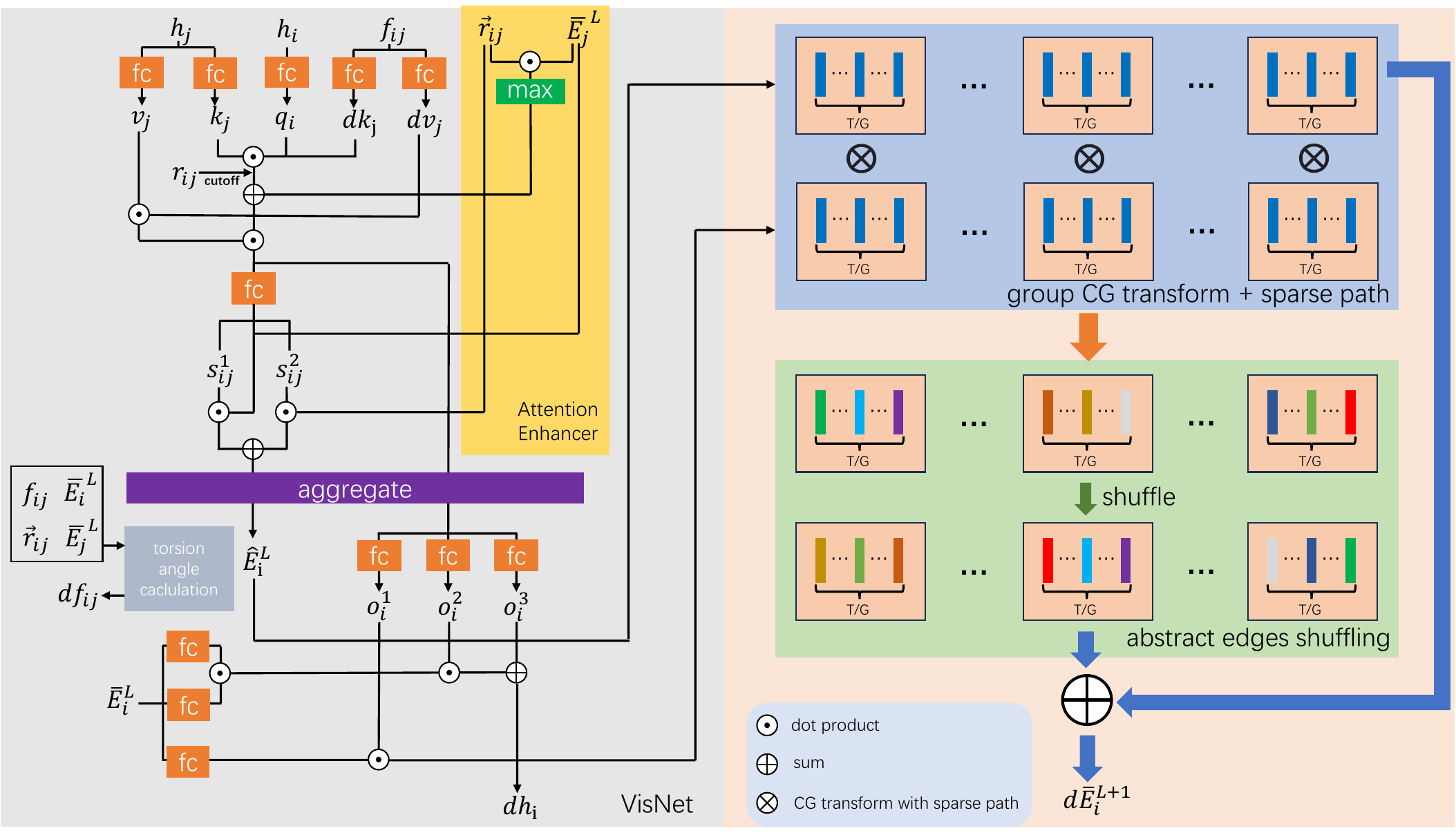}
  \vspace{-0.6cm}
  \caption{The architecture of a single layer of FreeCG. The cross-attention mechanism generates abstract edges through a permutation-invariant process. The abstract edges are also used to enhance the quality of the attention score, denoted as \textit{Attention Enhancer}. In the right part, the \textit{Group CG transform} organizes abstract edges into groups and performs the CG transform on each group. We adopt \textit{sparse path} for CG transform, enabling lower computation demands while maintaining O(3) equivariance. \textit{Abstract edges shuffling} improves the information exchange between different irreps. The details for sparse path and abstract edges shuffling can be referred to Fig. \ref{fig:spandaes}.
  }
  \label{fig:fig1}
\vspace{-0.4cm}
\end{figure*}

\vspace{-0.45cm}

\section{Methods}
\vspace{-0.4cm}

\subsection{Background}
\vspace{-0.25cm}

\textbf{Group, equivariance and invariance.} Permutation, rotation, and translation form different groups in group theory. Formally, a set with a binary operation $(G,*)$ is said to be a group if and only if the following conditions hold: 1) $g_1*g_2\in G,$ for any $g_1,g_2\in G$ (closure) 2) $(g_1*g_2)*g_3 = g_1 * (g_2*g_3),$ for any $g_1,g_2,g_3\in G$ (associativity) 3) There exists a group element $e\in G$, such that $g*e=e*g=g,$ for any $g\in G$. ($e$ identity element) 4) There is a group element $g^\prime$ \textit{w.r.t.} $g$, such that $g*g^\prime=g^\prime*g=e$, for each $g\in G$ ($g^\prime$ inverse element). The group elements $g\in G$, according to the representation theory, can be represented as linear transformations $\mathcal{P}_V(g)\in GL(V)$  on vector space $V$. Given a function $f:X\to Y$, where $X$ and $Y$ are vector spaces. It is said to be $G$-equivariant if and only if $f(\mathcal{P}_X(g) x)=\mathcal{P}_Y(g)f(x)$, for any $g\in G$. $G$-invariance is a special case when $\mathcal{P}_Y(g)$ is an identity matrix. Permutation equivariance and ${\text E}(3)$-equivariance are two properties each layer of our model obeys. Permutation equivariance means the index of node or edge features will be consistent when passing a layer. ${\text E}(3)$-equivariance covers rotation, translation, and reflection, where the translation is explicitly guaranteed via only considering the relative distances between atoms, thus we consider ${\text O}(3)$-equivariance where translations are omitted. It is intuitive to correspondingly change directional features when the whole molecule rotates or reflects.

\noindent\textbf{Tensor, irreps and CG transform.} Tensors are high-dimensional generalizations of scalars, vectors, and matrices. Scalars and vectors are both special cases of Cartesian tensors. Tensor product can generate high-rank tensors from low-rank ones. Formally, tensors are the results of tensor product of several vectors and covectors. In our context, it is not essential to distinguish between vectors and covectors. Tensors representing groups can be further decomposed to the direct sum of irreps. For example, tensors of ${\rm SO}(3)$ (omit reflection compared to ${\text O}(3)$) on 9-space (from tensor product of two $3\times 3$ rotation matrix) can be decomposed into $1\times 1$ ($l=0$), $3\times 3$ ($l=1$), and $5\times 5$ ($l=2$) irreps, which are called Wigner-D matrices. In EGNNs, we often project the distance vector between atoms onto the unit sphere $S^2$ with the central atom as the center of sphere. Actually, $S^2$ is homomorphic to the quotient group SO(3)$/$SO(2), thus it also has its own irreps, \textit{e.g.,} $l=0$ scalar and $l=1$ vector. $S^2$ irreps are the main features we maintain in our model, where irreps with degree $l$ has $2l+1$ elements, which are often indexed by $m$. To combine these features, we can calculate the tensor product between them, and the results can, again, be decomposed to irreps. This process is known as CG transform, which utilizes CG coefficients to perform transformations. For instance, $A^{1,l_2l_3\mapsto l_1}_{m_1}=\sum_{m_2,m_3} C^{l_1l_2l_3}_{m_1m_2m_3} A^{2,l_2}_{m_2} A^{3,l_3}_{m_3}$, where $A^{l}$ are $S^2$ irreps, $m$ denotes the elements of irreps, and $C$ the CG coefficient. To satisfy ${\text O}(3)$, we consider an additional variable, parity $p$, which takes the values of $1$ or $-1$. Irreps with $p=-1$ will be inverse when the space is reflected, and $p=1$ unchanged. The above formula of CG transform becomes:

\begin{equation}
\begin{aligned}
A^{1,l_2p_2l_3p_3\mapsto l_1p_1}_{m_1}=\mathbbm{1}_{(p_1=p_2p_3)}\sum_{m_2,m_3}  C^{l_1l_2l_3}_{m_1m_2m_3} 
A^{2,l_2p_2}_{m_2} A^{3,l_3p_3}_{m_3}
\end{aligned}
\end{equation}
where $\mathbbm{1}_{(expression)}$ is the indicator function, outputting $1$ if $expression$ is true, and $0$ otherwise. Given a vector ($l=1$ $S^2$ irreps), we can \textit{lift} it to irreps with arbitrary degree $l$ and $p=(-1)^l$, via a series of real spherical harmonics $(Y^l_{m=1},...,Y^l_{m=2l+1})$. For further details about group theory, we refer interested readers to related books and papers \citep{zee2016group,raczka1986theory,thomas2018tensor,jeevanjee2011introduction,cohen2018spherical}.

\begin{figure*}[tb]
  
  \centering
  \includegraphics[width=1.\linewidth]{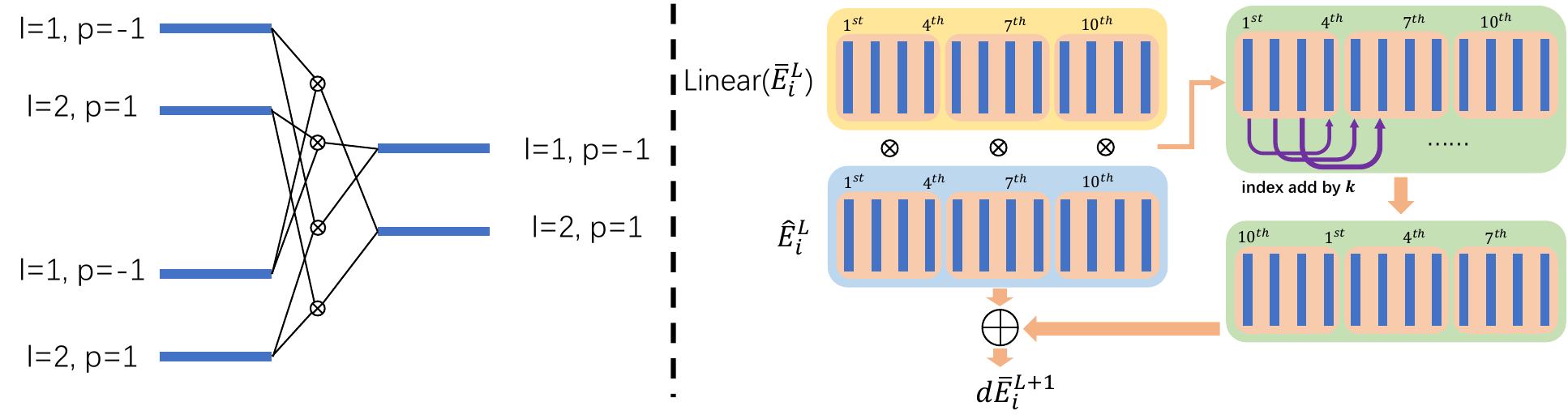}
  \vspace{-0.5cm}
  \caption{Details on sparse path and abstract edges shuffling. \textbf{Left:} The sparse path holds two useful properties: 1) The number of paths is less than the weaker SO(3) equivariance (4 vs. 8). 2) Each output irreps contains the information from input ones with both degree $l=1$ and $l=2$. \textbf{Right:} The shuffling strategy is to add a constant $k$ for the index of each abstract edge. The shuffled result is then added by $\hat{E}^L_i$, and get the final added value $d\overline{E}^{L+1}_i$.
  }
  \label{fig:spandaes}
\vspace{-0.4cm}
\end{figure*}

\vspace{-0.3cm}
\subsection{Problem analysis}\label{sec:framework}
\vspace{-0.2cm}
The task of force field prediction can be formalised as follows: Given a set of atoms with their positions and atom types $\{\bm{X}, \bm{Z}\}$, the neural network $f_\theta$ with parameter $\theta$ aims to predict the energy, and by which it derives the predicted force on each atom. In each layer of NequIP \citep{batzner20223}, messages from neighboring atoms are aggregated and combined with the features of the central atom. The messages are created via CG transform between the irreps. Here, we revisit the critical step constructing messages to a central atom $a$ in NequIP:

\begin{equation}
\label{eq:nequipl}
\begin{aligned}
    \mathcal{L}^{l_ep_el_np_n\mapsto l_op_o}_{acm_o}(\bm{X},\bm{N}) = \mathbbm{1}_{(p_o=p_ep_n)}
 \sum_{m_em_n} C^{l_ol_el_n}_{m_om_em_n} \\
 \sum_{b\in \mathcal{N}(a)} (R(\lVert \vec{r}_{ab}\rVert)^{l_ol_el_n}_{c}) Y^{l_e}_{m_e}(\frac{\vec{r}_{ab}}{\lVert \vec{r}_{ab}\rVert})N^{l_np_n}_{bcm_n}
\end{aligned}
\end{equation}
where $\mathcal{N}(a)$ is the set of neighboring atoms of atom $a$. $R$ is a MLP. $\lVert*\rVert$ is Euclidean norm. $N_b$ is the features of node $b$. $\vec{r}_{ab}$ is the vector pointing from atom $a$ to $b$. Consider the vector function form of Eq. \ref{eq:nequipl}:
$\bm{\mathcal{L}}^{l_ep_el_np_n\mapsto l_op_o}_{cm_o} = (\mathcal{L}^{l_ep_el_np_n\mapsto l_op_o}_{1cm_o},\mathcal{L}^{l_ep_el_np_n\mapsto l_op_o}_{2cm_o},...)$, which is permutation equivariant w.r.t. permutation operations acting on $\bm{X}$ and $\bm{N}$. Formally, it means $\bm{\mathcal{L}}^{l_ep_el_np_n\mapsto l_op_o}_{cm_o}(\mathcal{P}_{\bm{X}}\bm{X},\mathcal{P}_{\bm{N}}\bm{N}) = \mathcal{P}_{\bm{\mathcal{L}}}\bm{\mathcal{L}}^{l_ep_el_np_n\mapsto l_op_o}_{cm_o}(\bm{X},\bm{N})$. Put simply, if we exchange the \textit{indexes} of two atoms, for example, 1 and 2, and feed them into function $\bm{\mathcal{L}}^{l_ep_el_np_n\mapsto l_op_o}_{cm_o}$, it equals to that we directly change the index 1 and 2 of the output of function $\bm{\mathcal{L}}^{l_ep_el_np_n\mapsto l_op_o}_{cm_o}$, which is $(\mathcal{L}^{l_ep_el_np_n\mapsto l_op_o}_{2cm_o},\mathcal{L}^{l_ep_el_np_n\mapsto l_op_o}_{1cm_o},...)$. This property is simple and very important for the molecular neural networks, as the properties of a molecule should not depend on the order in which these atoms are arranged.

Most works take this property for granted. However, the permutation equivariance is actually important but vulnerable. It limits the design space to a very small scope, and make the network poorly scalable when the number of neighbors arises. Specifically, it brings the following issues:

\vspace{-0.25cm}
\begin{problem}
\label{problem1}
    The CG transform layer scales as $\mathcal{O}(\max\limits_i {\rm card}(\mathcal{N}(i)))$, where ${\rm card}(X)$ is the number of elements in set $X$. One cannot arbitrarily remove calculations for a specific neighboring atom because it would break the permutation equivariance. 
\end{problem}
\vspace{-0.3cm}
\begin{problem}
\label{problem2}
    The design space is limited for maintaining permutation equivariance. For example, in Eq. \ref{eq:nequipl}, the formulation and the parameters of $R$ should be the same across different neighboring atoms, thus forbidding the design for complicated CG transform layers.
\end{problem}
\vspace{-0.25cm}

Problem \ref{problem1} poses heavy computation challenges, as the CG transform itself is very time-consuming, compared to dot product and element-wise multiplication. We provide a detailed analysis for the efficiency of CG transform in Sec. \ref{sec:efficcg}. On the other hand, the narrowness for design space brought by problem \ref{problem2} makes it hard to design a high expressive CG transform layer, as only limited structures can be designed to maintain permutation equivariance. To address these problems, we aim to free the CG transform in messages transmissions from the constraints of permutation equivariance without compromising the overall equivariance of the network. Here, we leverage a simple and useful mathematical property. Consider a function $h$ that can be written as:

\begin{equation}
    h(x) = h^{'}(h_1(x),h_2(x),...)
\end{equation}
if $h_*(x)$ are all $G$-invariant, then, regardless of how we design $h^{'}$, the overall function $h$ must be $G$-invariant as well. The proof is simple, as:

\begin{equation}
\begin{aligned}
    h^{'}(h_1(\mathcal{P}_{X}(g)x),h_2(\mathcal{P}_{X}(g)x),...) =  \mathcal{P}_{h}(e)h^{'}(h_1(x),h_2(x),...)
\end{aligned}
\end{equation}
Here, the invariance we care about is w.r.t. the internal features of a given atoms and the neighbours. Specifically, it is the sum $\sum_{b\in \mathcal{N}(a)} (R(\lVert \vec{r}_{ab}\rVert)^{l_ol_el_n}_{c}) Y^{l_e}_{m_e}(\frac{\vec{r}_{ab}}{\lVert \vec{r}_{ab}\rVert})N^{l_np_n}_{bcm_n}$, and the term for each $b$. Invariant components will not affect the equivariance of the layer. Thus, we first obtain a set of invariant functions $h_{*}$. Then, we can freely design the function $h^{'}$ based on these invariant functions.

\vspace{-0.25cm}
\subsection{FreeCG}
\vspace{-0.2cm}
\label{sec:freecg}
\noindent\textbf{Abstract edges.} The above proposition presents an elegant way to solve problem \ref{problem1} and \ref{problem2}. The idea is to put CG transform inside the function $h^{'}$, and by the conclusion, we can completely free the design space of the CG transform. The first step is to construct the permutation invariant function $h_*$. To emphasize the geometric information, we want these $h_*$'s to be the aggregation of edge features. We call $h_*$'s \textit{abstract edges}. For the concrete design, we take the transformer architecture in ViSNet \citep{wang2024enhancing} as an efficient tool to construct abstract edges. The detailed information of the implementation is in the Sec. \ref{sec:modelimplementation}. In ViSNet, each edge maintains high-degree features $E_{ij}=E^{l=1}_{ij} \oplus E^{l=2}_{ij}$ consisting of irreps $E^{l}_{ij}=Y^l(\vec{r}_{ij}/\lVert \vec{r}_{ij} \rVert)$. The above features are invariant to layer index $L$. The computed attention $a^{L,t}_{ij}$ is multiplied to each edge. The sum of them $\hat{E}^{L}_{i,t} = \sum_{ij\in\mathcal{E}(i)} a^{L}_{ij,t}E^L_{ij}$ forms an temporary abstract edge, where we omit the degree $l$, and $L$ the index of the layer. $t$ denotes the index of the $t$-th abstract edges $(\hat{E}^{L}_{i,t=1},\hat{E}^{L}_{i,t=2},...)$.
In the original ViSNet, it was used to update the geometric feature $d\overline{E}^{L+1}_{i} = \hat{E}^{L}_i + o^{L,1}_i\cdot{\rm Linear}(\overline{E}^{L}_{i})$, where ${\rm Linear}$ is a fully-connected linear operation, which performs across the dimension of $t$, thus does not break the equivariance, and $o^{L,1}_i$ is a variable generated from the node feature, as we will introduce in Sec. \ref{sec:modelimplementation}. We leverage the fact that $\hat{E}^{L}$ fits the requirements of $h_*$ in our proposition, take them as abstract edges, and propose methods to construct CG transform function $h$ upon it. The proof that each abstract edge meets the requirement for $h_*$, namely, it is permutation invariant, is in Sec. \ref{sec:proofpermut}.

\vspace{-0.05cm}

\noindent\textbf{Group CG transform.} The number of abstract edges is decided by us, so the complexity for computing CG transform in Problem \ref{problem1} is controlled to be constant. The proposition above gives us enough freedom to construct the CG transform function $h$, expanding the design space to maximum, alleviating Problem \ref{problem2}. The idea is to use CG transform to replace the updating mechanism of $\overline{E}$ in ViSNet. A naïve attempt is to directly take the CG transform between $\overline{E}^L$ and $\hat{E}^L$ to acquire $\overline{E}^{L+1}$. However, we want to further decrease the $O(T^2)$ time complexity for the CG transform, where $T$ is the number of abstract edges, even though it is a constant number. Leveraging the unlimited freedom in constructing $h$, and taking inspiration of group convolution \citep{krizhevsky2012imagenet}, we propose group CG transform (distinct from the group in group theory). We first split the abstract edges of $\overline{E}^L$ and $\hat{E}^L$ into groups, where each index of abstract edge belongs to some group $U_g$ , the integer $g$ ranges from $1$ to $G$, and $G$ a hyper-parameter for the number of total groups. Then a group CG transform acts as:

\vspace{-0.1cm}

\begin{equation}
\label{eq:cgfully2}
\begin{aligned}
    d\overline{E}^{\prime L+1,l_o,p_o}_{i,t_om_o} = \mathbbm{1}_{(p_o=p_1p_2)} o^{L,1}\sum_{l_1,l_2}\sum_{m_1,m_2}C^{l_o,l_1,l_2}_{m_om_1m_2} \\
    \sum_{t_1,t_2\in U_g}W^{l_o,l_1,l_2}_{t_ot_1t_2} \rm{Linear}(\overline{E}^{L}_i)^{l_1,p_1}_{t_1m_1} \hat{E}^{L,l_2,p_2}_{i,t_2m_2}
\end{aligned}
\end{equation}

\vspace{-0.1cm}

where $t_o\in U_g$. The group CG transform decreases the time complexity to $O(T^2/G)$. Here, the parameters $W$ for CG transform are also worth emphasizing. They are not necessary to be kept the same across different abstract edges $t$ to keep permutation equivariance, and do not need to adopt the same MLP for each edge to calculate weights. Thus, we directly assign different weights $W$ for different abstract edges to enhance the model expressive ability. In contrast to previous methods, We save the computational cost for calculating weights for each edge.

\vspace{-0.05cm}

\noindent\textbf{Sparse path.} Typically, ensuring SO(3) equivariance is considered more efficent than ensuring O(3) equivariance. It is because we often need to consider both $p=1$ and $p=-1$ for a single $l$ for O(3), thus the total computation is quadrupled, and memory usage is doubled. Here we propose a method to keep O(3) while being more efficient than SO(3). We only keep $(l=1, p=-1)$ and $(l=2,p=1)$, which is same as the order of directly using spherical harmonics. In such way, It suffices that each output irreps containing information from both input irreps through CG transform, as $(l=1, p=-1)*(l=2,p=1)\mapsto(l=1,p=-1)$, $(l=1, p=-1)*(l=1,p=-1)\mapsto (l=2,p=1)$, $(l=2, p=1)*(l=2,p=1)\mapsto (l=2,p=1)$, and $(l=1, p=-1)*(l=1,p=-1)\mapsto (l=2,p=1)$. There are only 4 path in contrast to 8 path for SO(3), being O(3) equivariant but more efficient than being SO(3) equivariant, illustrated in Fig. \ref{fig:spandaes}.

\vspace{-0.05cm}

\noindent\textbf{Abstract edges shuffling.} Inspired by ShuffleNet \citep{zhang2018shufflenet}, we can also shuffle the abstract edges to make the information exchanged comprehensively. We shuffle all the abstract edges. Specifically, we increase the indices of all irreps by $1.5*T/G$. If the index exceeds $T$, we start counting from 1 again. Theoretically, the shuffling strategy can be arbitrary as long as maintaining the same strategy for each layer during every inference. This process is also depicted in Fig. \ref{fig:spandaes}. The ablation on different strategies is shown in Results section.

\vspace{-0.05cm}

\noindent\textbf{Abstract edges enhance cross-attention.} The Transformer integrates neighboring atoms information in molecular tasks through cross-attention mechanism, which aims to capture relations for those atoms exhibiting strong interatomic correlations. In order to better utilize abstract edge information, we leverage it to augment the generation of attention scores. To calculate the cross-attention, the node scalar features are processed to generate query $Q$, key $K$, and value $V$ for each atom, respectively. Then, the self attention is computed as $A_{ij} = \mathop{\rm SiLU}(Q_i\odot K_j)$, where $\odot$ represents dot product, and SiLU is the activation function. Note that ViSNet is different from other transformer-based models where $A_{ij}$ is scaled by the SiLU instead of Softmax across different $j$. We integrate the information of abstract edges by:

\vspace{-0.1cm}

\begin{equation}
\label{eq:attnenhancer}
    A_{ij} = \mathop{\rm SiLU}(Q_i\odot K_j + \max\limits_t(\overline{E}^{L}_{j,t}\odot E_{ij}) )
\end{equation}

\vspace{-0.1cm}

where $\max \limits t( \overline { E } ^ { L } _ { j,t } \odot E { ij } )$ denotes the maximum value of the dot product between each abstract edge $\overline { E } ^ { L } _ { j,t }$ and the real edge features $E_ { ij }$ across different abstract edges. $E_ { ij }$ does not have an $L$ superscript because these features remain constant across different layers. This as an additional contribution to the cross-attention, as it quantifies how well the abstract edges capture the information of the edge linking atoms $i$ and $j$. The generation of abstract edges and the operations performed on them are illustrated in Fig. \ref{fig:fig1}, respectively.

\begin{table*}[]
\setlength{\tabcolsep}{5pt}
\renewcommand{\arraystretch}{1.2}
\resizebox{\linewidth}{!}{\begin{tabular}{@{}lccccccccccc@{}}
\toprule
Molecule        & SchNet & DimeNet & PaiNN & SpookeyNet & ET             & GemNet & NequIP & SO3KRATES & ViSNet         & QuinNet & FreeCG         \\ \midrule
\multicolumn{12}{c}{\textit{Energy Prediction}}                                                                                                    \\ \midrule
Aspirin         & 0.37   & 0.204   & 0.167 & 0.151      & 0.123          & -      & 0.131  & 0.139     & 0.116          & 0.119   & \textbf{0.110} \\
Ethanol         & 0.08   & 0.064   & 0.064 & 0.052      & 0.052          & -      & 0.051  & 0.052     & 0.051          & 0.050   & \textbf{0.049} \\
Malondialdehyde & 0.13   & 0.104   & 0.091 & 0.079      & 0.077          & -      & 0.076  & 0.077     & \textbf{0.075} & 0.078   & 0.094          \\
Naphthalene     & 0.16   & 0.122   & 0.116 & 0.116      & 0.085          & -      & 0.113  & 0.115     & 0.085          & 0.101   & \textbf{0.083} \\
Salicylic acid  & 0.20   & 0.134   & 0.116 & 0.114      & 0.093          & -      & 0.106  & 0.016     & 0.092          & 0.101   & \textbf{0.090} \\
Toluene         & 0.12   & 0.102   & 0.095 & 0.094      & \textbf{0.074} & -      & 0.092  & 0.095     & \textbf{0.074} & 0.080   & 0.076          \\
Uracil          & 0.14   & 0.115   & 0.106 & 0.105      & \textbf{0.095} & -      & 0.104  & 0.103     & \textbf{0.095} & 0.096   & 0.097          \\ \midrule
\multicolumn{12}{c}{\textit{Force Prediction}}                                                                                                     \\ \midrule
Aspirin         & 1.35   & 0.499   & 0.338 & 0.258      & 0.253          & 0.217  & 0.184  & 0.236     & 0.155          & 0.145   & \textbf{0.122} \\
Ethanol         & 0.39   & 0.230   & 0.224 & 0.094      & 0.109          & 0.085  & 0.071  & 0.096     & 0.060          & 0.060   & \textbf{0.053} \\
Malondialdehyde & 0.66   & 0.383   & 0.319 & 0.167      & 0.169          & 0.155  & 0.129  & 0.147     & 0.100          & 0.097   & \textbf{0.095} \\
Naphthalene     & 0.58   & 0.215   & 0.077 & 0.089      & 0.061          & 0.051  & 0.039  & 0.074     & 0.039          & 0.039   & \textbf{0.034} \\
Salicylic acid  & 0.85   & 0.374   & 0.195 & 0.180      & 0.129          & 0.125  & 0.090  & 0.145     & 0.084          & 0.080   & \textbf{0.070} \\
Toluene         & 0.57   & 0.216   & 0.094 & 0.087      & 0.067          & 0.060  & 0.046  & 0.073     & 0.039          & 0.039   & \textbf{0.035} \\
Uracil          & 0.56   & 0.301   & 0.139 & 0.119      & 0.095          & 0.097  & 0.076  & 0.111     & 0.062          & 0.062   & \textbf{0.059} \\ \bottomrule
\end{tabular}}
\vspace{-0.2cm}
\caption{
    \noindent Performances on MD17 dataset. The results are reported in mean abosolute error (MAE). The energies and forces are measured in kcal/mol and kcal/mol/$\angstrom$, respectively. The best numbers are marked in \textbf{bold}.}
\label{tab:md17}
\vspace{-0.4cm}

\end{table*}

\begin{table*}[]

\setlength{\tabcolsep}{5pt}
\renewcommand{\arraystretch}{1.2}
\resizebox{\linewidth}{!}{\begin{tabular}{@{}lccccccccc@{}}
\toprule
Molecule       & UNiTE & GemNet & NequIP          & MACE            & Allergo         & BOTNet          & ViSNet          & QuinNet & FreeCG          \\ \midrule
\multicolumn{10}{c}{Energy Prediction}                                                                                                                \\ \midrule
Aspirin        & 0.055 & -      & 0.0530          & 0.0507          & 0.0530          & 0.0530          & \textbf{0.0445} & 0.0486  & 0.0530          \\
Azobenzene     & 0.025 & -      & 0.0161          & 0.0277          & 0.0277          & 0.0161          & \textbf{0.0156} & 0.0394  & 0.0217          \\
Benzene        & 0.002 & -      & 0.0009          & 0.0092          & 0.0069          & \textbf{0.0007} & \textbf{0.0007} & 0.0096  & 0.0107          \\
Ethanol        & 0.014 & -      & 0.0092          & \textbf{0.0032} & 0.0092          & 0.0092          & 0.0078          & 0.0096  & 0.0087          \\
Malonaldehyde  & 0.025 & -      & 0.0184          & 0.0185          & 0.0138          & 0.185           & \textbf{0.0132} & 0.0168  & 0.0146          \\
Naphthalene    & 0.011 & -      & \textbf{0.0046} & 0.1153          & \textbf{0.0046} & \textbf{0.0046} & 0.0057          & 0.0174  & 0.0118          \\
Paracetamol    & 0.044 & -      & 0.0323          & 0.0300          & 0.0346          & 0.0300          & \textbf{0.0258} & 0.0362  & 0.0392          \\
Salicylic acid & 0.017 & -      & \textbf{0.0161} & 0.0208          & 0.0208          & 0.0185          & \textbf{0.0161} & 0.033   & 0.0233          \\
Toluene        & 0.010 & -      & 0.0069          & 0.0115          & 0.0092          & 0.0069          & \textbf{0.0059} & 0.0139  & 0.0334          \\
Uracil         & 0.013 & -      & 0.0092          & 0.0115          & 0.0138          & 0.0092          & \textbf{0.0069} & 0.0149  & 0.0116          \\ \midrule
\multicolumn{10}{c}{Force Prediction}                                                                                                                 \\ \midrule
Aspirin        & 0.175 & 0.2191 & 0.1891          & 0.1522          & 0.1684          & 0.1900          & 0.1520          & 0.1429  & \textbf{0.1212} \\
Azobenzene     & 0.097 & -      & 0.0669          & 0.0692          & 0.0600          & 0.0761          & 0.0585          & 0.0513  & \textbf{0.0486} \\
Benzene        & 0.017 & 0.0115 & 0.0069          & 0.0069          & \textbf{0.0046} & 0.0069          & 0.0056          & 0.0047  & 0.0056          \\
Ethanol        & 0.085 & 0.083  & 0.0646          & 0.0484          & 0.0484          & 0.0738          & 0.0522          & 0.0516  & \textbf{0.0438} \\
Malonaldehyde  & 0.152 & 0.1522 & 0.0118          & 0.0946          & 0.0830          & 0.1338          & 0.0893          & 0.0875  & \textbf{0.0802} \\
Naphthalene    & 0.060 & 0.0438 & 0.0300          & 0.0369          & \textbf{0.0208} & 0.0415          & 0.0291          & 0.0242  & 0.0228          \\
Paracetamol    & 0.164 & -      & 0.1361          & 0.1107          & 0.1130          & 0.1338          & 0.1029          & 0.0979  & \textbf{0.0840} \\
Salicylic acid & 0.088 & 0.1222 & 0.0922          & 0.0715          & 0.0669          & 0.0992          & 0.0795          & 0.0771  & \textbf{0.0648} \\
Toluene        & 0.058 & 0.0507 & 0.0369          & 0.0350          & 0.0415          & 0.0438          & 0.0264          & 0.0244  & \textbf{0.0239} \\
Uracil         & 0.088 & 0.0876 & 0.0669          & 0.0484          & \textbf{0.0415} & 0.0738          & 0.0495          & 0.0487  & 0.0446          \\ \bottomrule
\end{tabular}}
\vspace{-0.2cm}
\caption{Performances on rMD17 dataset. The results are reported in MAE. The energies and forces are measured in kcal/mol and kcal/mol/$\angstrom$, respectively. The best numbers are marked in \textbf{bold}.}
\vspace{-0.4cm}
\label{tab:rmd17}
\end{table*}

\vspace{-0.45cm}
\section{Experiments}
\vspace{-0.35cm}
To evaluate the performance of FreeCG, we collect molecular force field datasets, on which we compare our methods with other SOTA MLFFs, which include small molecules dataset MD17 \citep{chmiela2017machine} with its revised version, rMD17 \citep{Christensen2020}, and large molecules dataset MD22 \citep{chmiela2023accurate}. To test the generalization capacity of the proposed FreeCG, we also evaluate the performance of FreeCG on a standard molecule property prediction dataset, QM9 \citep{ruddigkeit2012enumeration,ramakrishnan2014quantum}. We take popular SOTA models into the comparison, including sGDML \citep{chmiela2017machine}, SchNet \citep{schutt2018schnet}, DimeNet \citep{gasteiger2020directional}, SphereNet \citep{liu2021spherical}, PaxNet \citep{zhang2022efficient}, PaiNN \citep{schutt2021equivariant}, SpookyNet \citep{unke2021spookynet}, ForceNet \citep{hu2021forcenet}, ET \citep{tholke2021equivariant}, GemNet \citep{gasteiger2021gemnet}, ComENet \citep{wang2022comenet}, NequIP \citep{batzner20223}, UniTE \citep{qiao2022informing}, SO3KRATES \citep{frank2022so3krates}, MACE \citep{batatia2022mace}, Allegro \citep{musaelian2023learning}, BOTNet \citep{batatia2022design}, ViSNet \citep{wang2024enhancing}, ViSNet-LSRM \citep{li2023long}, and QuinNet \citep{wang2023quinnet}. To asses the practicality of FreeCG on real-world tasks, Molecular dynamics simulations are run for MD17 molecules and two periodic systems, water \citep{fu2022forces,wu2006water} and LiPS \citep{batzner20223} under PBCs. FreeCG is also evaluated on the conformation space of a 166-atom mini-protein, Chignolin \citep{wang2023aimd}. The results reveal that FreeCG is capable to make accurate predictions on force and energy efficiently, and it also exhibits strong practicality on real-world applications. The ablation on each component and corresponding hyper-parameters of FreeCG are also presented. The common settings and extra experiments are reported in the Sec. \ref{sec:commonset}.

\vspace{-0.5cm}

\subsection{Comparison with state-of-the-arts for MLFFs}
\vspace{-0.3cm}

\noindent\textbf{Force field on small molecules and periodic systems.} MD17 is a famous molecular dynamics benchmark for small molecules. FreeCG outperforms others in all force prediction tasks. It also significantly decreases the force prediction errors for the most hard-to-predict molecule in this datasets, aspirin, by 15\%. Remarkably, FreeCG also decreases the MAE by over 10\% for ethanol, naphthalene, and salicylic acid. FreeCG does not have a particular preference for the size of molecules. It demonstrates strong performance for aspirin (180.2 g/mol) and excels on ethanol (46.1 g/mol). The energy prediction is also competitive when compared to other SOTA methods. rMD17 is the revised version of MD17. It recomputed the trajectories of each atom with higher accuracy. The force prediction accuracy of FreeCG is still leading in majority of the molecules. It improves the force results compared to the baseline model, ViSNet, in all molecules except for the benzene, and performs SOTA on more atoms. Note that the results on benzene is already extreme high with previous models. The results for MD17 and rMD17 can be referred to Tab. \ref{tab:md17} and \ref{tab:rmd17}, respectively. The accuracy of force prediction on periodic systems, water and LiPS, are also evaluated. Fig. \ref{fig:mdwater} and \ref{fig:mdlips} show that, FreeCG achieves the best performance across other methods, and decrease the MAE value of the second best by 50\% for water.

\noindent\textbf{Force field on large molecules.} MD22 is a large molecules benchmark adopted by several studies \citep{wang2024enhancing,wang2023quinnet,li2023long}. As shown in Tab. \ref{tab:md22}, it reveals that FreeCG also performs well for large scale data. It leads in most tracks for force prediction, and shows comparable results for energy prediction. Remarkably, The decreasing in MAE for energy and force prediction on Ac-Ala3-NHMe are both around 20\%. The performances for the other models are not consistent well for force prediction, while ViSNet-LSRM exhibits strong performance for energy prediction. It is also reasonable that all modern deep neural network-based methods outperform sGDML, as a classical kernel method.

\begin{table*}[]
\centering

\setlength{\tabcolsep}{5pt}
\renewcommand{\arraystretch}{1.2}
\resizebox{\linewidth}{!}{\begin{tabular}{@{}lccccccc@{}}
\toprule
Molecule               & \multicolumn{1}{l}{sGDML} & \multicolumn{1}{l}{ViSNet} & \multicolumn{1}{l}{ViSNet-Improper} & \multicolumn{1}{l}{ViSNet-LSRM} & \multicolumn{1}{l}{MACE} & \multicolumn{1}{l}{QuinNet} & \multicolumn{1}{l}{FreeCG} \\ \midrule
\multicolumn{8}{c}{\textit{Energy Prediction}}                                                                                                                                                                                                \\ \midrule
Ac-Ala3-NHMe           & 0.391                     & 0.0636                     & 0.0546                              & 0.0673                          & 0.0631                   & 0.0840                      & \textbf{0.0507}             \\
AT-AT                  & 0.720                     & 0.0708                     & 0.0668                              & 0.0780                          & 0.108                    & 0.144                       & \textbf{0.0665}            \\
AT-AT-CG-CG            & 1.42                      & 0.196                      & 0.197                               & \textbf{0.118}                  & 0.154                    & 0.379                       & 0.254                      \\
DHA                    & 1.29                      & 0.0741                     & \textbf{0.0700}                     & 0.0897                          & 0.135                    & 0.118                       & 0.0761                     \\
Buckyball catcher      & 1.17                      & 0.508                      & 0.537                               & \textbf{0.319}                  & 0.489                    & 0.563                       & 0.512                      \\
Stachyose              & 4.00                      & 0.0915                     & 0.0882                              & \textbf{0.104}                  & 0.122                    & 0.226                       & 0.183                      \\
Double-walled nanotube & 4.00                      & 0.800                      & 0.601                               & 1.81                            & 1.67                     & 1.81                        & \textbf{0.543}             \\ \midrule
\multicolumn{8}{c}{\textit{Force Prediction}}                                                                                                                                                                                                 \\ \midrule
Ac-Ala3-NHMe           & 0.790                     & 0.0830                     & 0.0709                              & 0.0942                          & 0.0876                   & 0.0681                      & \textbf{0.0531}            \\
AT-AT                  & 0.690                     & 0.0812                     & 0.0776                              & 0.0781                          & 0.0992                   & 0.0687                      & \textbf{0.0634}            \\
AT-AT-CG-CG            & 0.700                     & 0.148                      & 0.139                               & \textbf{0.1064}                 & 0.1153                   & 0.1273                      & 0.1252                     \\
DHA                    & 0.750                     & 0.0598                     & 0.0554                              & 0.0598                          & 0.0646                   & 0.0515                      & \textbf{0.0507}            \\
Buckyball catcher      & 0.680                     & 0.184                      & 0.201                               & 0.1026                          & \textbf{0.0853}          & 0.1091                      & 0.1783                     \\
Stachyose              & 0.680                     & 0.0879                     & 0.0802                              & 0.0767                          & 0.0876                   & \textbf{0.0543}             & 0.612                      \\
Double-walled nanotube & 0.520                     & 0.362                      & 0.292                               & 0.3391                          & 0.2767                   & 0.2473                      & \textbf{0.2449}            \\ \bottomrule
\end{tabular}}
\vspace{-0.2cm}
\caption{Performances on MD22 dataset. The results are reported in MAE. The energies and forces are measured in kcal/mol and kcal/mol/$\angstrom$, respectively. The best numbers are marked in \textbf{bold}. Note that the energy MAE is calculated without being divided by the total number of atoms as \cite{wang2024enhancing}, unlike \cite{wang2023quinnet,chmiela2017machine,chmiela2023accurate}, which does not affect the comparison.}
\vspace{-0.6cm}
    \label{tab:md22}
\end{table*}

\vspace{-0.5cm}

\subsection{Comparison with state-of-the-arts for molecular properties predictions}
\vspace{-0.3cm}

% \noindent\textbf{Molecular property prediction.} 
To examine the generalization power on molecular property prediction of FreeCG, we collect QM9 as a standard benchmark for this task. FreeCG performs the best for most properties. ViSNet also performs the second best in most measures. Although these two methods are proposed to be MLFFs, but they are even more comparable than others in molecular property prediction tasks. The results are in Tab. \ref{tab:qm9}.

\begin{figure*}[tb]

  \centering
  \includegraphics[width=0.75\linewidth]{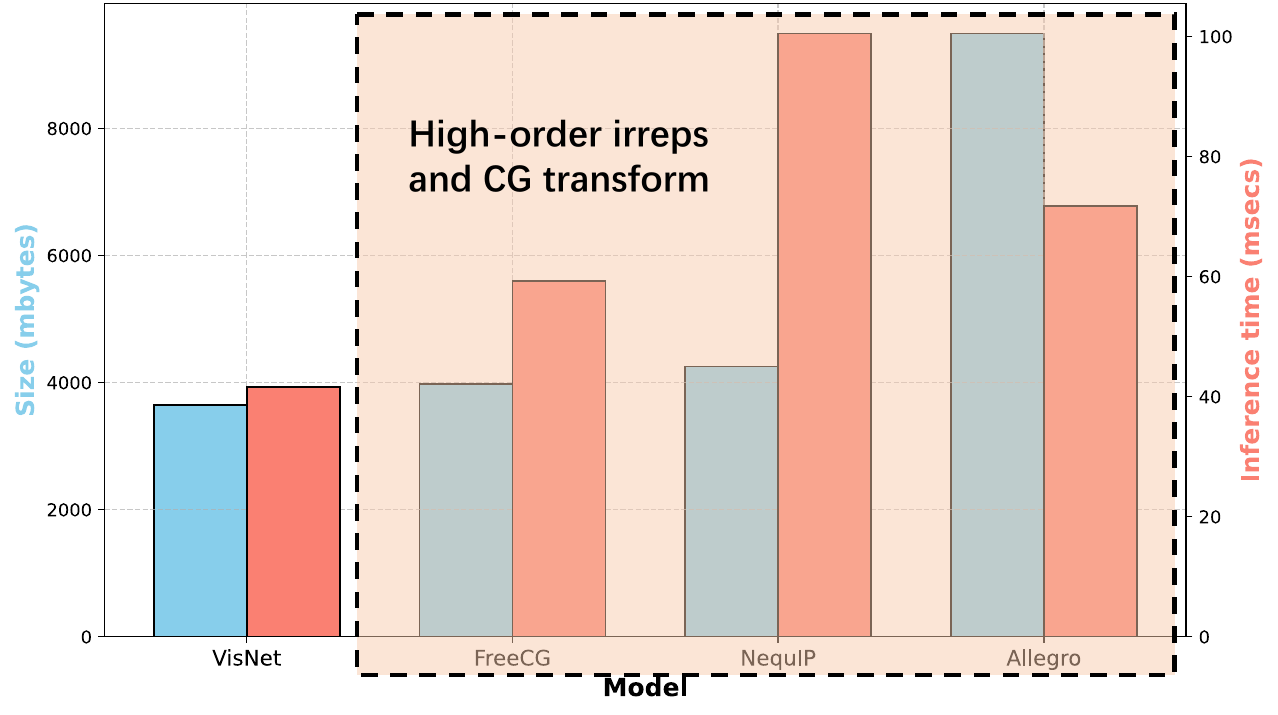}
  \vspace{-0.4cm}
  \caption{The speed and memory occupation of FreeCG compared with other SoTA models. Numbers are reported based on a single chignolin molecule. The right three models are based on high-order irreps and CG transform.
  }
  \label{fig:benchmarkchignolin}
\vspace{-0.5cm}
\end{figure*}

\vspace{-0.4cm}

\subsection{Efficiency benchmarking}
\vspace{-0.2cm}

%\noindent\textbf{Efficiency benchmarking.} 
AIMD-Chig dataset is taken as a benchmark for testing the memory usage and inference speed. We compare the inference speed and memory usage of FreeCG with ViSNet, NequIP, and Allegro. The results are shown in Fig. \ref{fig:benchmarkchignolin}. FreeCG adds little extra time and memory cost, compared to the baseline model, ViSNet. It is also the most efficient one for both memory and speed, compared to the other two CG transform-based methods, NequIP and Allegro. The overall results prove the effciency of FreeCG. The number of groups in group CG transform also impacts the inference speed. Fig. \ref{fig:groupvstime} shows the theoretical number of paths and the actual inference time for different group numbers. A computation analysis for CG transform can be referred to Sec. \ref{sec:efficcg}.

\vspace{-0.3cm}

\subsection{Ablation study}
\vspace{-0.2cm}

We conduct ablations on different modules we propose, as well as the strategies for abstract edges shuffling. The results are shown in Tab. \ref{tab:ablation}. It reveals that each of our module contributes to the final score of FreeCG. In the final implementation of abstract edges shuffling, we add the index of each abstract edge by $1.5*T/G$. Here we also study the influence of the shuffling strategies. We adopt $0.5*T/G$, $1.0*T/G$, and $1.5*T/G$ for comparing the performance. We can see from the result that $1.5*T/G$ works the best. The group numbers are also evaluated and a large number of groups appears to be a good choice.

\begin{table*}[]

\setlength{\tabcolsep}{5pt}
\renewcommand{\arraystretch}{1.2}
\resizebox{\linewidth}{!}{\begin{tabular}{llcccccccccc}
\hline
Target                &            & SchNet & EGNN & DimeNet++ & PaiNN & SphereNet & PaxNet & ET   & ComENet & ViSNet        & FreeCG        \\ \hline
$\mu$                 & mD         & 33     & 29   & 29.7      & 12    & 24.5      & 10.8   & 11   & 24.5    & \textbf{9.5}  & 11.4          \\
$\alpha$              & m$a^3_0$   & 235    & 71   & 43.5      & 45    & 44.9      & 44.7   & 59   & 45.2    & 41.1          & \textbf{38.2} \\
$\epsilon_{HOMO}$     & meV        & 41     & 29   & 24.6      & 27.6  & 22.8      & 22.8   & 20.3 & 23.1    & 17.3          & \textbf{16.6} \\
$\epsilon_{LUMO}$     & meV        & 34     & 25   & 19.5      & 20.4  & 18.9      & 19.2   & 17.5 & 19.8    & 14.8          & \textbf{13.5} \\
$\Delta \epsilon$     & meV        & 63     & 48   & 32.6      & 45.7  & 31.1      & 31     & 36.1 & 32.4    & 31.7          & \textbf{31.5} \\
$\langle R^2 \rangle$ & m$a^2_0$   & 73     & 106  & 331       & 66    & 268       & 93     & 33   & 259     & \textbf{29.8} & 82.1          \\
$ZPVE$                & meV        & 1.7    & 1.55 & 1.21      & 1.28  & 1.12      & 1.17   & 1.84 & 1.2     & 1.56          & \textbf{1.10} \\
$U_0$                 & meV        & 14     & 11   & 6.32      & 5.85  & 6.26      & 5.9    & 6.15 & 6.59    & 4.23          & \textbf{4.11} \\
$U$                   & meV        & 19     & 12   & 6.28      & 5.83  & 6.36      & 5.92   & 6.38 & 6.82    & \textbf{4.25} & 4.51          \\
$H$                   & meV        & 14     & 12   & 6.53      & 5.98  & 6.33      & 6.04   & 6.16 & 6.86    & 4.52          & \textbf{4.13} \\
$G$                   & meV        & 14     & 12   & 7.56      & 7.35  & 7.78      & 7.14   & 7.62 & 7.98    & 5.86          & \textbf{5.65} \\
$C_v$                 & mcal/mol K & 33     & 31   & 23        & 24    & 22        & 23.1   & 26   & 24      & 23            & \textbf{20.4} \\ \hline
\end{tabular}}
\vspace{-0.2cm}
\caption{Molecular property prediction on QM9 dataset. The results are reported in MAE. The best numbers are marked in \textbf{bold}.}
\vspace{-0.2cm}
\label{tab:qm9}
\end{table*}

\begin{table*}[]
\centering

\resizebox{0.5\linewidth}{!}{\begin{tabular}{@{}lccc@{}}
\toprule
\multirow{2}{*}{Method} & \multicolumn{3}{c}{Aspirin}                       \\ \cmidrule(l){2-4} 
                        & Val loss        & Energy         & Force          \\ \midrule
ViSNet                  & -               & 0.116          & 0.155          \\ \midrule
+ Group CG transform  &                 &                &                \\
8 groups               & 0.0509          & 0.123          & 0.144          \\
32 groups                & 0.0416          & 0.112          & 0.129          \\ \midrule
+ Abstract edges shuffling       &                 &                &                \\
1-group shuffle         & 0.0401          & 0.112          & 0.128          \\
0.5-group shuffle       & 0.0396          & 0.110          & 0.128          \\
1.5-group shuffle       & 0.0384          & 0.111          & 0.125          \\ \midrule
+ Attention enhancer    & \textbf{0.0345} & \textbf{0.110} & \textbf{0.122} \\ \bottomrule
\end{tabular}}
\vspace{-0.2cm}
\caption{Ablation on different modules. Abstract edges shuffling and Attention enhancer are added upon the best choices of the above modules, with respect to the validation loss.}
\label{tab:ablation}
\vspace{-0.5cm}
\end{table*}

\vspace{-0.2cm}

\subsection{Extension to other models}
\vspace{-0.2cm}
% \noindent\textbf{Extension to other models.} 
The free CG transform proposed in this work also presents a paradigm to enhance other geometric models. To demonstrate that the FreeCG concept is applicable to other models, we use the former SOTA model, QuinNet, as an example to illustrate how effectively FreeCG can be extended to other architectures. QuinNet has a transformer architecture inside, so we construct abstract edges the same way FreeCG does, and adopts all FreeCG modules upon those abstract edges. We train both the vanilla QuinNet and QuinNet equipped with FreeCG modules for 1000 epochs, evaluating on the test set every 200 epochs. As shown in Fig. \ref{fig:quinnetforce} and \ref{fig:quinnetenergy}, when equipped with FreeCG, QuinNet significantly gets improved for both energy and force prediction. This trend becomes more pronounced with longer training periods, as evidenced by the results at the 1000\textit{th} epoch.

\vspace{-0.4cm}

\section{Conclusion}
\vspace{-0.2cm}
This work proposes FreeCG, an equivariant neural network that frees the design space of CG transform. It achieves SOTA performance in force prediction for MD17, rMD17, and MD22 datasets, as well as in molecular properties prediction for QM9, with only minor computational overhead. The practicality of FreeCG for conducting molecular dynamics simulations is thoroughly examined across periodic systems, small molecules in MD17, and the mini-protein Chignolin. As we show FreeCG helps improve QuinNet, it also introduces a new paradigm for expressive and efficient CG transform-based neural network design in the future.

\bibliography{iclr2025_conference}
\bibliographystyle{iclr2025_conference}

\clearpage

\appendix
\section{Appendix}

\subsection{Experimental settings}
\label{sec:commonset}
We conduct all the experiments under the same software and hardware settings. The machine is equipped with an Intel$^\circledR$ Xeon$^\circledR$ Gold 6330 CPU @ 2.00GHz, with NVIDIA Tesla A100 80G GPUs. We run the experiments for each molecule on a single GPU. Pytorch 1.10.0 is used as the basic machine learning python library. For the CG transform operations, we adopt e3nn 0.5.1. Matplotlib 3.0.3 is utilized for plotting. The details can be referred to Tab. \ref{tab:settings}. We report the hyperparameters used in Tab. \ref{tab:hyperparam}. For training/validation/test splits, we follow previous works \citep{wang2024enhancing,wang2023quinnet,fu2022forces}. We pick up the model for evaluating on test set based on the performance on the validation set. If the model does not improve for a given number of epochs, we will terminate the training and select the checkpoint with the best validation score. As previous works, Exponential Moving Average (EMA) is adopted to generate the model weights. The detailed training configurations are shown in Tab. \ref{tab:hyperparam}.

\begin{table*}[htbp]

\resizebox{\linewidth}{!}{\begin{tabular}{@{}ccccc@{}}
\toprule
\multicolumn{2}{c}{Hardware}                                                                                              & \multicolumn{3}{c}{Software}               \\ \midrule
CPU                                                                                                   & GPU               & Neural Network & Equivariance & Plotting   \\ \midrule
\begin{tabular}[c]{@{}c@{}}Intel$^\circledR$ Xeon$^\circledR$ \\ Gold 6330 CPU @ 2.00GHz\end{tabular} & NVIDIA Tesla A100 & Pytorch 1.10.0       & e3nn 0.5.1         & Matplotlib 3.0.3 \\ \bottomrule
\end{tabular}}
\caption{Hardware and software settings.}
\label{tab:settings}
\end{table*}

\begin{table*}[htbp]
\centering

\resizebox{1.0\linewidth}{!}{\begin{tabular}{@{}lccccccc@{}}
\toprule
Hyperparameter               & MD17                    & rMD17                   & MD22                    & QM9     & Water-1k & LiPS & Chignolin             \\ \midrule
Initial learning rate        & 4e-4, 2e-4              & 2e-4                    & 2e-4, 1e-4              & 1e-4  & 5e-4 & 1e-3    & 2e-4             \\
Learning rate decay factor   & \multicolumn{7}{c}{0.8}                                                                               \\
Learning rate decay patience & 30                      & 30                      & 30                      & 15 & 5 & 5  & 10               \\
Learning rate warmup step    & 1000                    & 1000                    & 1000                    & 10000 & 1000 & 1000 & 1000                 \\
Optimizer                    & \multicolumn{7}{c}{AdamW ($\beta (0.9, 0.999)$)}                                                      \\
Epoch                        & 3000                    & 3000                    & 3000                    & 1500       & 1000 & 100 & 3000            \\
Batch size                   & 4                       & 4                       & 4                       & 32    & 1 & 1  & 4                \\
Number of layers             & 9 & 9 & 9 & 9 & 9 & 9 & 6                                                                              \\
Cutoff                       & 5.0, 4.0                & 5.0                     & 5.0, 4.0                & 5.0  & 6.0 & 6.0 & 5.0                  \\
Force/Energy loss weights    & 0.95/0.05               & 0.95/0.05               & 0.95/0.05               & -  & 1.0/0 & 1.0/0 & 0.95/0.05                    \\
Dimension of latent feature  & \multicolumn{1}{c}{256} & \multicolumn{1}{c}{256} & \multicolumn{1}{c}{256} & 512 & 256 & 256 & 128 \\ 
Number of groups             & \multicolumn{7}{c}{8}                                                                                 \\
Output head  & \multicolumn{7}{c}{Equivariant/Scalar} \\

EMA rate  & \multicolumn{7}{c}{0.999}

\\\bottomrule
\end{tabular}}
\caption{Hyperparameters for each dataset.}
\label{tab:hyperparam}
\end{table*}
\subsection{Model implementation}
\label{sec:modelimplementation}

Here we show how FreeCG is built upon ViSNet. This section provides detailed explanations of the implementation details, ensuring FreeCG can be replicated effectively.

\noindent\textbf{Input layer.} Given the atom coordinates and types $\{\bm{X}=\vec{r}_0, \vec{r}_1, \vec{r}_2,..., \vec{r}_N),\bm{Z}=(z_1,z_2,...,z_n)\}$, where $\vec{r}\in \mathbbm{R}^3$ the Cartesian coordinates of atom, and $z$ the atom type (atom numbers). First we embed the atom types to the latent space, and take them as our first layer's node features $h_i={\rm embedding}(z_i)\in \mathbbm{R}^C$. $C$ is the dimension of the latent space. For each atom, we only consider neighboring atoms within a given radius $\mathcal{N}(i)$, where we maintain the distance vector from the central atom to the neighboring atoms, and \textit{lift} them to $(l=1,p=-1)$ and $(l=2,p=1)$ irreps $E_{ij}\in \mathbbm{R}^{3+5}$ via real spherical harmonics applied on the unit vector $E_{ij}=Y^{l=1}(\vec{r}_{ij}/\lVert \vec{r}_{ij} \rVert)\oplus Y^{l=2}(\vec{r}_{ij}/\lVert \vec{r}_{ij} \rVert)$, where we also calculate the corresponding Euclidean norm $\lVert \vec{r}_{ij} \rVert$. The Euclidean norms of vectors are then converted to high-dimension scalar features (edge attributes) $f_{ij}={\rm RBF}(\vec{r}_{ij})\in\mathbbm{R}^C$ by radial basis functions (RBFs). We also maintain zero-initialized abstract edges $\overline{E}^{L=0}_i={\bm 0}$ for each node to be updated in the following layers. We assign the same number of abstract edges as the dimension of the latent features, such that additional operations to align the dimension numbers are not required.

\noindent\textbf{Intermediate layers.} Here, we use a superscript $L$ to denote the index of layer that the features are in. The message-passing between atoms is implemented by a transformer architecture. For each atom $i$, the neighboring atoms $j\in\mathcal{N}(i)$ will send messages to $i$, and the messages are aggregated to update the information of $i$. The query, key, and value of the node features are first calculated, respectively: $q_i=f_q(h_i)$, $k_j=f_k(h_j)$, $v_j=f_v(h_j)$. The edge attributes are also converted to auxiliary terms $dk_j=f_{dk}(f_{ij})$ and $dv_j = f_{dv}(f_{ij})$ to modulate keys and and values of atoms. Here functions $f$ are all fully-connected linear operations. Then we calculate the cross-attention between $i$ and $j$, which is 
\begin{equation}
\begin{aligned}
    a_{ij}={\rm SiLU}\bigg({\rm Cutoff}(\lVert\vec{r}_{ij}\rVert)q_ik_jdk_j +\\ {\rm AttEnhancer}(\vec{r}_{ij},\overline{E}^L_j)\bigg)
\end{aligned}
\end{equation}
where ${\rm Cutoff}(\cdot)$ is a cosine cutoff function, and ${\rm AttEnhancer}(\cdot)$ the proposed attention enhancer module, as we will formulate its details. First, recall the dimension of $\overline{E}^L_i\in \mathbbm{R}^{C*8}$ and $\vec{r}_{ij}\in\mathbbm{R}^{8}$. Each of the $C$ abstract edges will undergo a dot product with $\vec{r}_{ij}$. The highest value among them will be the output of ${\rm AttEnhancer}$. In other word,
\begin{equation}
{\rm AttEnhancer}(\overline{E}^L_i,\vec{r}_{ij})=\max_C(\overline{E}^L_i\odot \vec{r}_{ij})
\end{equation}
as we introduce in Eq. \ref{eq:attnenhancer}. Then the values are multiplied with $dv$ and attention. 
\begin{equation}
\hat{v}_{j\mapsto i}^L= v_j\cdot dv_j \cdot a_{ij}
\end{equation}
It then undergoes two different fully-connected operations to generate two coefficients $s_1$ and $s_2$. They are used to generate the abstract edges:
\begin{equation}
\hat{E}^L_{j\mapsto i}=\overline{E}^L_i \cdot s_1 + E_{ij} \cdot s_2
\end{equation}
This variable, together with $\hat{v}_{j\mapsto i}$, are aggregated by sum:
\begin{equation}
\hat{E}^L_{i}=\sum_{j\in\mathcal{N}(i)}\hat{E}^L_{j\mapsto i}
\end{equation}
\begin{equation}
\hat{v}^L_{i}=\sum_{j\in\mathcal{N}(i)}\hat{v}^L_{j\mapsto i}
\end{equation}
$\hat{v}^L_{i}$ then converts to three variables for further operation: 
\begin{equation}
o^{L,1}_i,o^{L,2}_i,o^{L,3}_i={\rm Linear}(\hat{v}^L_i)
\end{equation}
$\hat{E}^L_{i}\in\mathbbm{R}^{C*8}$ and $\overline{E}^L_{i}\in\mathbbm{R}^{C*8}$ are used for the following group CG transform and abstract edges shuffling. First $\overline{E}^L_{i}\in\mathbbm{R}^{C*8}$ undergoes a fully-connected operation along $C$ dimension, and multiply with $o^{L,1}_i$, which means we get $o^{L,1}_i\cdot{\rm Linear}(\overline{E}^L_{i})$.
It, together with $\hat{E}^L_{i}\in\mathbbm{R}^{C*8}$, are then divided into $G$ groups along $C$ dimension, where we get $\hat{E}^{L}_{i,t\in G_g}\in \mathbbm{R}^{\frac{C}{G}*8}$, and $(o^{L,1}_i \cdot {\rm Linear}(\overline{E}^{L}_{i}))_{t\in G_g}\in \mathbbm{R}^{\frac{C}{G}*8}$. Then, we perform CG transform between two variables in a fully connected form with learnable weights, and concatenate the results to generate $d\overline{E}^{\prime L+1}_i$ before shuffling, as shown in Eq. \ref{eq:cgfully2}. For the shuffling strategies, we add $\frac{3C}{2G}$ to each index of the abstract edges $\overline{E}^{\prime L+1}_i$. Then, it is added with $\hat{E}^{L}_i$ to form a residual structure, as we show here:
\begin{equation}
    d\overline{E}^{L+1}_i={\rm shuffle}(d\overline{E}^{\prime L+1}_i) + \hat{E}^{L}_i
\end{equation}
where $d\overline{E}^{L+1}_i$ is added to $\overline{E}^{L}_i$ to obtain $\overline{E}^{L+1}_i$. Next, we update $h$ and $f$. We first show the update for $h$:
\begin{equation}
\begin{aligned}
    dh^{L+1}_{i}= h^{L}_{i} + \bigg({\rm Linear_1}(\overline{E}^L_i)\odot {\rm Linear_2}(\overline{E}^L_i)\bigg)\cdot o^{L,2}_i \\ + o^{L,3}_i
\end{aligned}
\end{equation}
To update $f$, we follow ViSNet to leverage rejection of vectors:
\begin{equation}
\begin{aligned}
    df^{L+1}_{ij}= f^{L}_{ij} + {\rm RejCalc_{trg}}(\overline{E}^L_i,\vec{r}_{ij})\odot \\ {\rm RejCalc_{src}} (\overline{E}^L_i,\vec{r}_{ij}) \cdot {\rm SiLU}({\rm Linear}(f^L_{ij}))
\end{aligned}
\end{equation}
where rejection calculation module $RejCalc$ is:
\begin{equation}
    RejCalc_{mode}(a,b)=a - ({\rm Linear_{mode}}(a)\odot b)\cdot b
\end{equation}
The updated $\overline{E}^{L+1}$, $h^{L+1}$, and $f^{L+1}$ are fed into the next layer.

\noindent\textbf{Output layers} are different with respect to the task our model performs on. We introduce the details for each task.

\noindent\textit{Force field prediction.} FreeCG is based on energy-conservative field, which means we derive the force from the predicted potential energy. Following ViSNet \citep{wang2024enhancing} and PaiNN \citep{schutt2021equivariant}, we predict the potential energy of the molecule via equivariant gated module.
\begin{equation}
\label{eq:equihead1}
h^{L+1}_i, u^{L+1}_i = {\rm MLP}\bigg({\rm Concat}(h_i^{L},\lVert{\rm Linear_1}(\overline{E}^{L}_i)\rVert)\bigg)
\end{equation}
where ${\rm MLP}$ is an $1$-hidden layer multi-layer preceptor. There is one more step to update $\overline{E}^{L+1}_i$:
\begin{equation}
\label{eq:equihead2}
\overline{E}^{L+1}_i = {\rm Linear_2}(\overline{E}^{L}_i) \cdot u^{L+1}_i
\end{equation}
These calculations are then repeated twice in succession. There is also an alternative head design for force field prediction which uses only scalar features $h$. Under this setting, Eq. \ref{eq:equihead1} and (\ref{eq:equihead2}) are replaced by:
\begin{equation}
h^{\bm L} = {\rm MLP}(h^{{\bm L}-1})
\end{equation}
In Tab. \ref{tab:hyperparam} where we introduce our hyper-parameter choices, the scalar version of output head is denoted as \textit{Scalar}, and the other one \textit{Equivariant}. Finally, the total energy of the molecule is the sum of the last-layer node features $h_i^{\bm L}\in \mathbbm{R}$:
\begin{equation}
y = \sum_i h^{\bm L}_i
\end{equation}
and the force is the negative gradients of the total energy:
\begin{equation}
F_i = -\nabla_{\vec{r}_i} y
\end{equation}

\noindent\textit{Property prediction.} The calculations for properties in QM9 follow the same procedure as energy prediction in force field prediction, with the exception of molecular dipole and electronic spatial extent. We first need to calculate the center of mass $\vec{r}_c$, which is:
\begin{equation}
\vec{r}_c = \frac{\sum_i m_i\cdot \vec{r}_i}{\sum_i m_i}
\end{equation}
For molecular dipole, the formula is:
\begin{equation}
\mu = \left\Vert\sum_i\overline{E}_i^{\bm L} + h_i^{\bm L}(\vec{r}_i-\vec{r}_c) \right\Vert
\end{equation}
and for electronic spatial extent:
\begin{equation}
    \langle R^2 \rangle = \sum_i h_i^{\bm L}\lVert \vec{r}_i-\vec{r}_c\rVert
\end{equation}
The output head can be easily adapted for different tasks, providing flexibility in property prediction.

\begin{figure*}[tb]

  \centering
  \includegraphics[width=1.\linewidth]{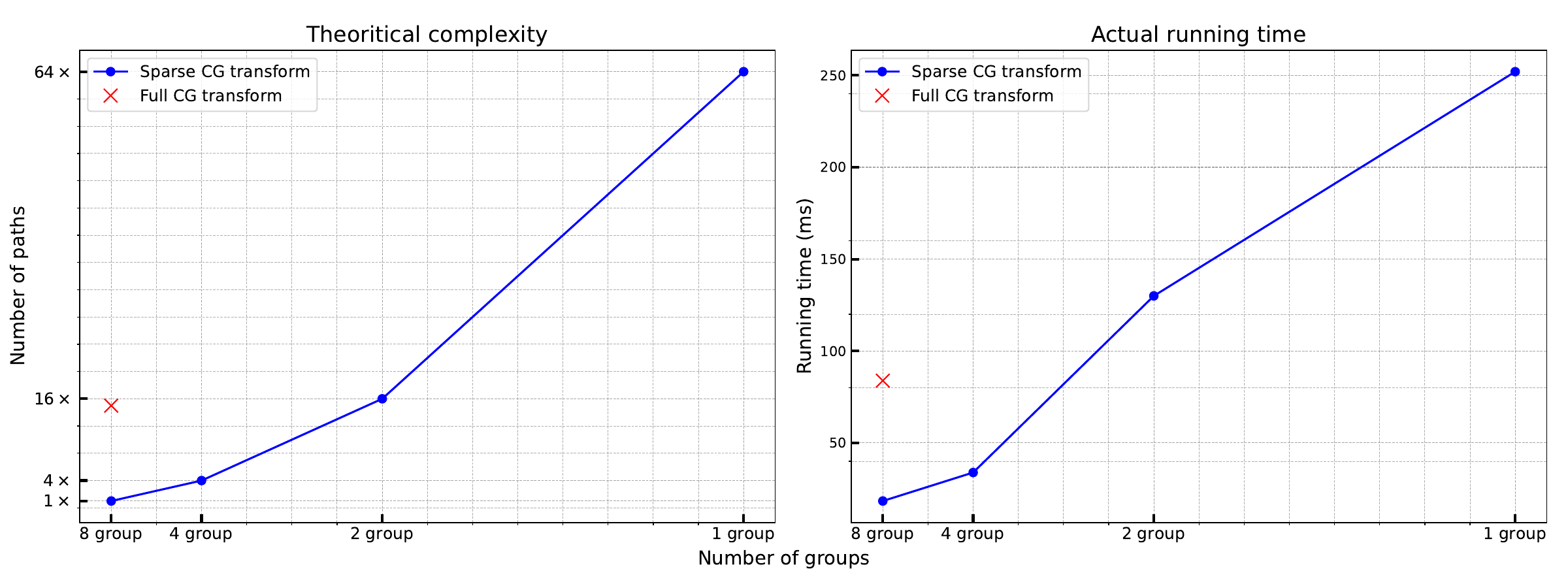}
  
  \caption{Efficiency analysis of group CG transform. \textbf{Left:} The number of paths for CG transform under different group numbers, where the numbers of irreps are the same. \textbf{Right:} The actual running time for CG transform for different group numers. Here we adopt sparse path strategy for computing 512 irreps (before grouping) for each $l$. Full CG transform denotes not using sparse path.
  }
  \label{fig:groupvstime}
\vspace{-0.2cm}
\end{figure*}

\begin{table*}[]
\label{tab:tpcomp}

\centering
\setlength{\tabcolsep}{5pt}
\renewcommand{\arraystretch}{1.2}
\resizebox{1.0\linewidth}{!}{\begin{tabular}{@{}cccccccccccccc@{}}
\toprule
\textbf{$l_o=2$} & 0                         & 1                         & 2 &  & \textbf{$l_o=1$} & 0                         & 1                         & 2  &  & \textbf{$l_o=2$} & 0 & 1  & 2  \\ \midrule
0                & \cellcolor[HTML]{34CDF9}1 &                           &   &  & 0                & -                         & \cellcolor[HTML]{34CDF9}3 & -  &  & 0                & - & -  & 5  \\
1                &                           & \cellcolor[HTML]{34CDF9}3 &   &  & 1                & \cellcolor[HTML]{34CDF9}3 & 6                         & 9  &  & 1                & - & 9  & 12 \\
2                &                           &                           & 5 &  & 2                & -                         & 9                         & 12 &  & 2                & 5 & 12 & 19 \\ \bottomrule
\end{tabular}}
\caption{The basic operation number for each type of CG transform. $l_o$ denotes the output degree. The column and row numbers represent the degrees of two input irreps, respectively. The cyan blocks represent the operations in regular neural networks, while the others are for high-order CG transform.}

\label{tab:tpcomplexity}
\end{table*}

\subsection{Datasets details}

\textbf{MD17 and rMD17.} They are both molecular dynamics datasets for small molecules. MD17 \citep{chmiela2017machine}, proposed by Chmiela, S., et al. contains \textit{ab-initio} level molecular dynamics trajectories. Four types of data are included in the dataset: atomic numbers, atomic positions, molecular energy, and the force acting on each atom. To alleivate the noise during the trajectory computation, Christensen, A. S. et al. also propose revised MD17 (rMD17) \citep{Christensen2020}, where molecular trajectories are calculated at the PBE/def2-SVP level of theory. The precision of the calculated trajectories is upheld by the tight SCF convergence and dense DFT integration grid. 

\noindent\textbf{MD22} consists of larger molecules with atoms numbering from 42 to 370, in contrast to MD17 and rMD17. The trajectories are sampled between 400K and 500K at 1fs resolution. The energy and force labels are obtained at the PBE+MBD level of theory. The root mean squared test error of force prediction is controlled to be around 1 kcal/mol/$\angstrom$ in the original paper \citep{chmiela2023accurate}. Thus, the training data sizes for different molecules vary. Generally, the larger the molecules, the smaller the training data size.

\noindent\textbf{QM9} consists of around 130,000 molecules with 12 properties regression tasks. It is a subset of the GDB-17 database \citep{ruddigkeit2012enumeration}. The data is calculated at B3LYP/6-31G(2df,p) based DFT level of accuracy. Since the attributes vary for different properties, we use distinct output head for each, as discussed in Sec. \ref{sec:modelimplementation}.

\noindent\textbf{Chignolin.} The AIMD-Chig dataset \citep{wang2023aimd} comprises of two million conformations of the 166-atom protein chignolin, obtained through sampling at the M06-2X/6-31 G* based DFT level. There are approximately 10,000 different conformations, including folded, unfolded, and metastable states. We report the performances of FreeCG on different parts of the energy landscape, and adopt this dataset to benchmark efficiency, following \citep{wang2024enhancing}.

\noindent\textbf{Periodic systems.} Periodic Boundary Conditions (PBCs) are vital in molecular dynamics simulations of periodic systems as they eliminate surface effects, enhance statistical sampling, and provide a realistic representation of bulk properties. Here we focus on two typical molecules, water and LiPS. The water dataset are generated by the flexible version of the Extended Simple Point Charge water model (SPC/E-fw) \citep{wu2006water} in \citep{fu2022forces}. The authors provides with several sizes of training sets, including -1k, -10k, 90k. Here we adopt the 1k version, where 950 samples are for training and the rest for validation. LiPS is an important solid-state materials for battery development. It can help predict key performance metrics such as capacity, energy density, and cycle life, aiding in the development of next-generation lithium-ion batteries. We follow the same train/val split as \citep{fu2022forces}. Some 3D structures of the data can be referred to Fig. \ref{fig_moleculestructure}.

\begin{figure*}[tb]
  
  \centering
  \includegraphics[width=1.\linewidth]{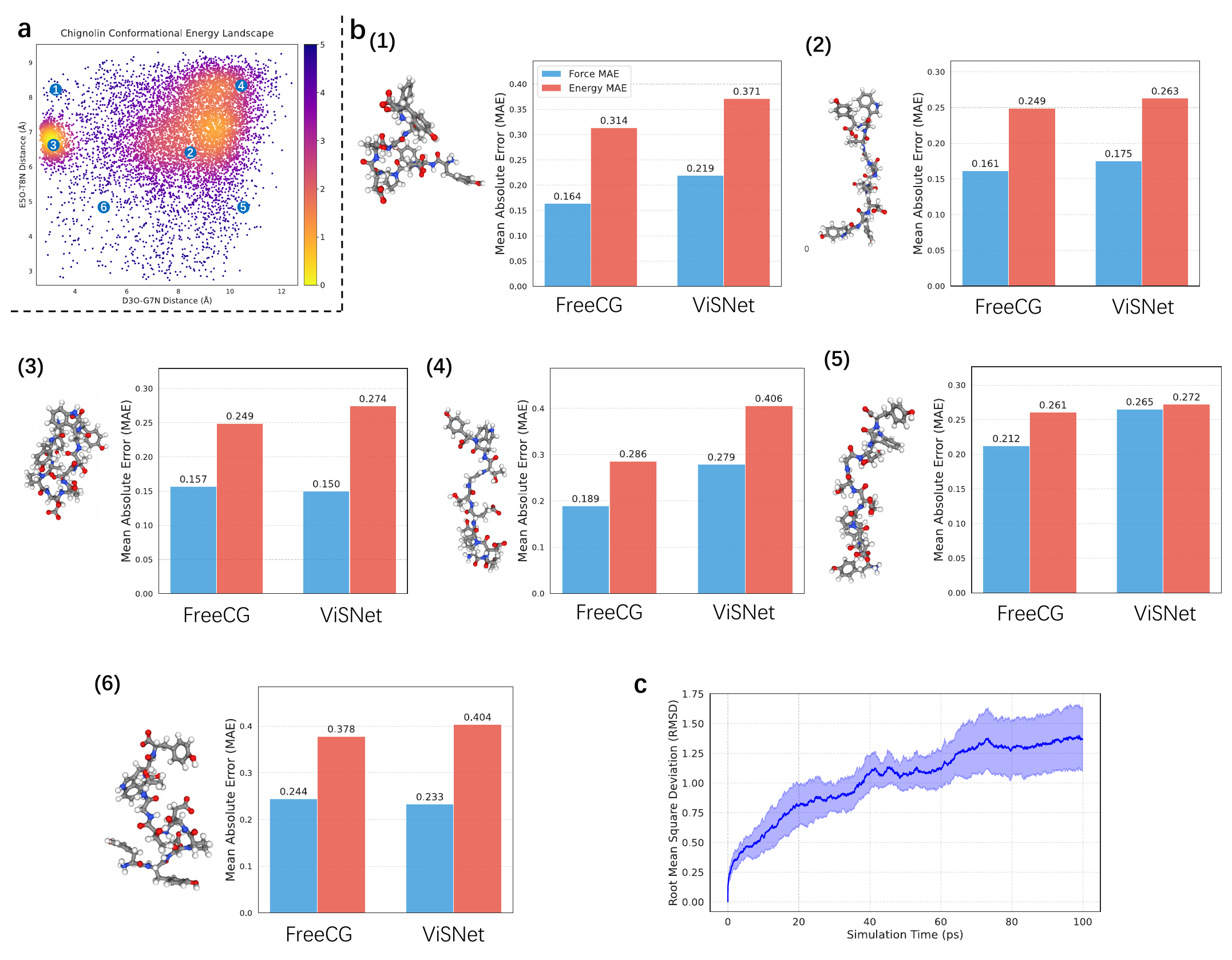}
  
  \caption{Applications for the 166-atom mini-protein, Chignolin. \textbf{a.} The energy landscape of Chignolin was sampled using Replica Exchange Molecular Dynamics (REMD). This landscape is characterized by two key distance parameters: the x-axis represents the distance between the carbonyl oxygen on the D3 backbone and the nitrogen on the G7 backbone, while the y-axis depicts the distance between the carbonyl oxygen on the E5 backbone and the nitrogen on the T8 backbone. These two distance metrics collectively illustrate the conformational states of Chignolin across its energy landscape. The left and right energy basins are corresponded to folded and unfolded states, respectively. \textbf{b.} The six conformations are sampled at the localization highlighted in the energy landscape. The force and energy performances (kcal/mol) are reported, with a comparison made to ViSNet. These six conformations cover both folded and unfolded states. \textbf{c.} The RMSD ($\si{\angstrom}$) during the molecular dynamics simulation. The shaded area denotes the values of standard derivations. The RMSD values are obtained by taking average of 10 trajectories.
  }
  \label{fig:chig}
\vspace{-0.2cm}
\end{figure*}

\subsection{Proof of the permutation invariance of abstract edges}
\label{sec:proofpermut}

According to Sec. \ref{sec:modelimplementation}, we first recall the last step for generating abstract edges:

\begin{equation}
    \hat{E}^L_{i}=\sum_{j\in\mathcal{N}(i)}\hat{E}^L_{j\mapsto i} = \sum_{p\in P} \sum_{j\in\mathcal{N}(i)}\frac{\mathcal{P}(p)\hat{E}^L_{j\mapsto i}}{{\rm Card}(P)}
\end{equation}
where $P$ is the set for all permutation operations, here we omit the subscript of $\mathcal{P}$ for specific spaces to work on. Here, note we are proving that the abstract edges for each atom are permutation invariant, and we can freely design CG transform \textit{per atom}, thus the permutation is applied to $j$ but not $i$.  It sums over all the permutation operations, and thus \textit{the last step} is permutation invariant. Then, it suffices to show that each of the previous step are all at least permutation equivariant. It also suffices to show they are permutation equivariant \textit{w.r.t.} single index switch operation, as each permutation operation can be made by several switches. If we exchange, without loss of generality, index $x$ and $y$, then those $a_{ij}$ that $x$ or $y$ shows up in the subscript for $j$ will exchange with each other, and so do $\hat{v}_{j\mapsto i}$ and $\hat{E}^L_{j\mapsto i}$. Thus, the rest steps are equivariant \textit{w.r.t.} single switch, and so they are permutation equivariant. Therefore, we conclude our proof that abstract edges are permutation invariant.

\subsection{Analysis on the efficiency of CG transform}
\label{sec:efficcg}
The CG transform consists of two steps: 1) performing a tensor product between two irreps, and 2) decomposing the resulting tensors into irreps. These transforms are actually quadratic homogeneous polynomials. For the sake of convenience, we discuss SO(3) group here. Recall the CG transform formula:

\begin{equation}
    D^{l_al_b\mapsto l_d}_{m_d} = \sum_{m_a,m_b} C^{l_dl_al_b}_{m_dm_am_b} A^{l_a}_{m_a} B^{l_b}_{m_b}
\end{equation}
where $m_a+m_b=m_d$. To illustrate, if we regard single multiplication and addition as the two basic operations, then combining two $l=1$ irreps to form a $l=2$ irreps will use up 1 basic operation for $m_d=\pm 2$, 3 operations for $m_d=\pm 1$, and 5 operations for $m_d=0$, making a total of 13 basic operations. One effective approach to understanding irreps is to view them as an extension of vectors and scalars. The dot product between vectors requires only 5 basic operations, in contrast to the 13 operations mentioned earlier. Hence, the CG transform is extremely time-consuming. The table of basic operations for the CG transform between each pair is shown in Tab. \ref{tab:tpcomplexity}.

\begin{figure*}[tb]

  \centering
  \includegraphics[width=1.\linewidth]{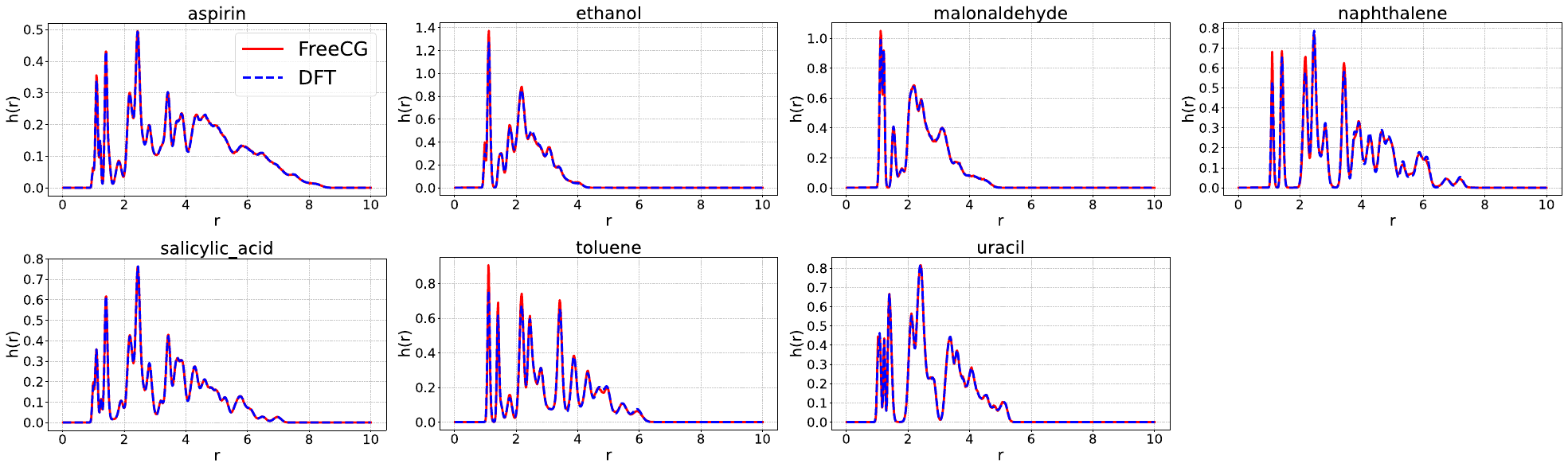}
  
  \caption{The distributions of interatomic distances $r$ during molecular dynamic simulations of MD17. The unit of $r$ is $\si{\angstrom}$, and the unit of $h(r)$ is $\si{\angstrom}^{-1}$.
  }
  \label{fig:md}
\vspace{-0.2cm}
\end{figure*}

\subsection{Additional experiments}

\noindent\textbf{Applications for full-atom proteins.} AIMD-Chig dataset \citep{wang2023aimd} comprises nearly 10,000 conformations of the 166-atom mini protein, Chignolin. These conformations were obtained including folded, unfolded, and metastable states. It is important to evaluate the performance of FreeCG on such real-world proteins. Following ViSNet \citep{wang2024enhancing}, as shown in Fig. \ref{fig:chig}, we explore the energy landscape of Chignolin, where we sample six conformations located at different parts of the landscape, covering folded and unfolded states. The energy and force predictions on these six conformations are compared with ViSNet. FreeCG succeeds in outperforming ViSNet for the most sampled conformations. Molecular dynamics simulations are run from six different initial conformations. To assess the simulation stability, we calculate the Root Mean Square Deviation (RMSD) between each step in the trajectory and the initial conformation, shown in Fig. \ref{fig:chig}(\textbf{c}). The results demonstrate satisfying performance of FreeCG on real-world proteins.

\noindent\textbf{Applications for molecular dynamics simulation.} We conduct molecular dynamics simulations for FreeCG on MD17 and compare the results with DFT calculations. We run a 300ps simulation for each molecule. The time step is set to 0.5 fs, under a Nosé–Hoover thermostat at 500K temperature. Like previous works \citep{fu2022forces,wang2024enhancing,wang2023quinnet}, we are interested in the distribution of interatomic distances. Here, $h(r)$ is defined as the probability density function of interatomic distances $r$. We plot the distribution as $h(r)$ \textit{w.r.t.} $r$, where $h(r)$ are averaged along frames or predicted trajectories, and we desire the distribution is similar to the MD17 datasets calculated by DFT. The results are shown in Fig. \ref{fig:md}, which shows that FreeCG is capable to well recover the interatomic distances distribution. We also conduct molecular dynamics simulation on water (1k version) \citep{fu2022forces,wu2006water} and LiPS \citep{batzner20223} datasets under periodic boundary conditions (PBCs) to evaluate how FreeCG performs on large molecular systems. We set the timestep to 0.25 fs and run total 200,000 steps for each type of molecules. We focus on the recovery of radial distribution functions (RDFs) because they effectively describe structural and thermodynamic properties. It is similarly calculated as $h(r)$ but under different constant multipliers. FreeCG finishes all the dynamics simulation with accurate recovery of the atomic distributions. The results for molecules under PBCs are shown in Fig. \ref{fig:mdwater} and \ref{fig:mdlips}.

\begin{figure*}[tb]

  \centering
  \includegraphics[width=1.\linewidth]{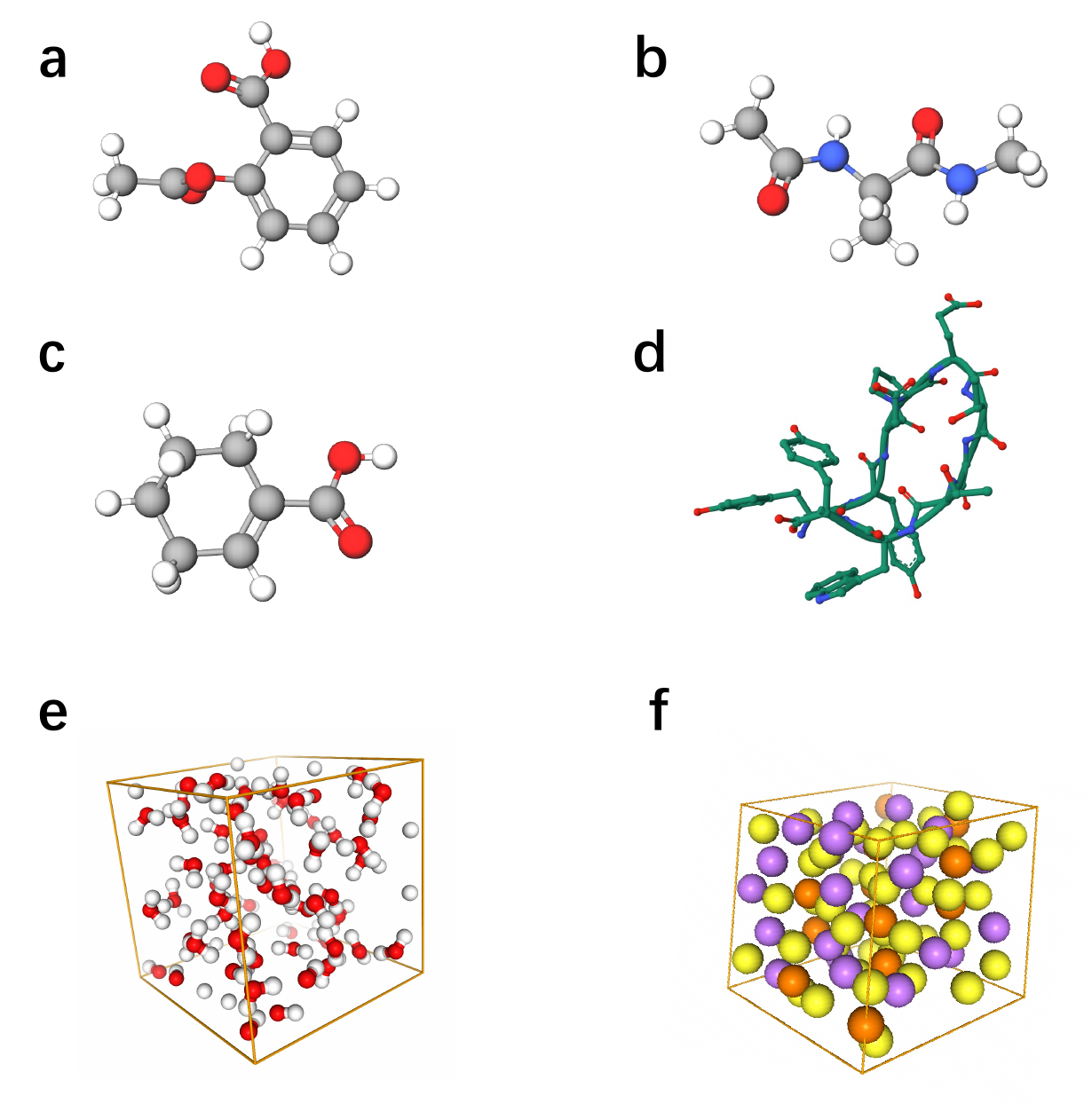}
  
  \caption{The 3D structures of the data considered in this work. \textbf{a.} Aspirin in MD17 and rMD17. \textbf{b}. Ac-Ala3-NHMe in MD22. \textbf{c.} 1-Cyclohexene-1-carboxylic acid in QM9. \textbf{d.} Chignolin. \textbf{e.} A single cell of water molecules under PBCs. \textbf{f.} A single cell of LiPS under PBCs (Note that the cell is not cubic).
  }
\vspace{-0.2cm}
\label{fig_moleculestructure}
\end{figure*}

\begin{figure*}[tb]

  \centering
  \includegraphics[width=1.\linewidth]{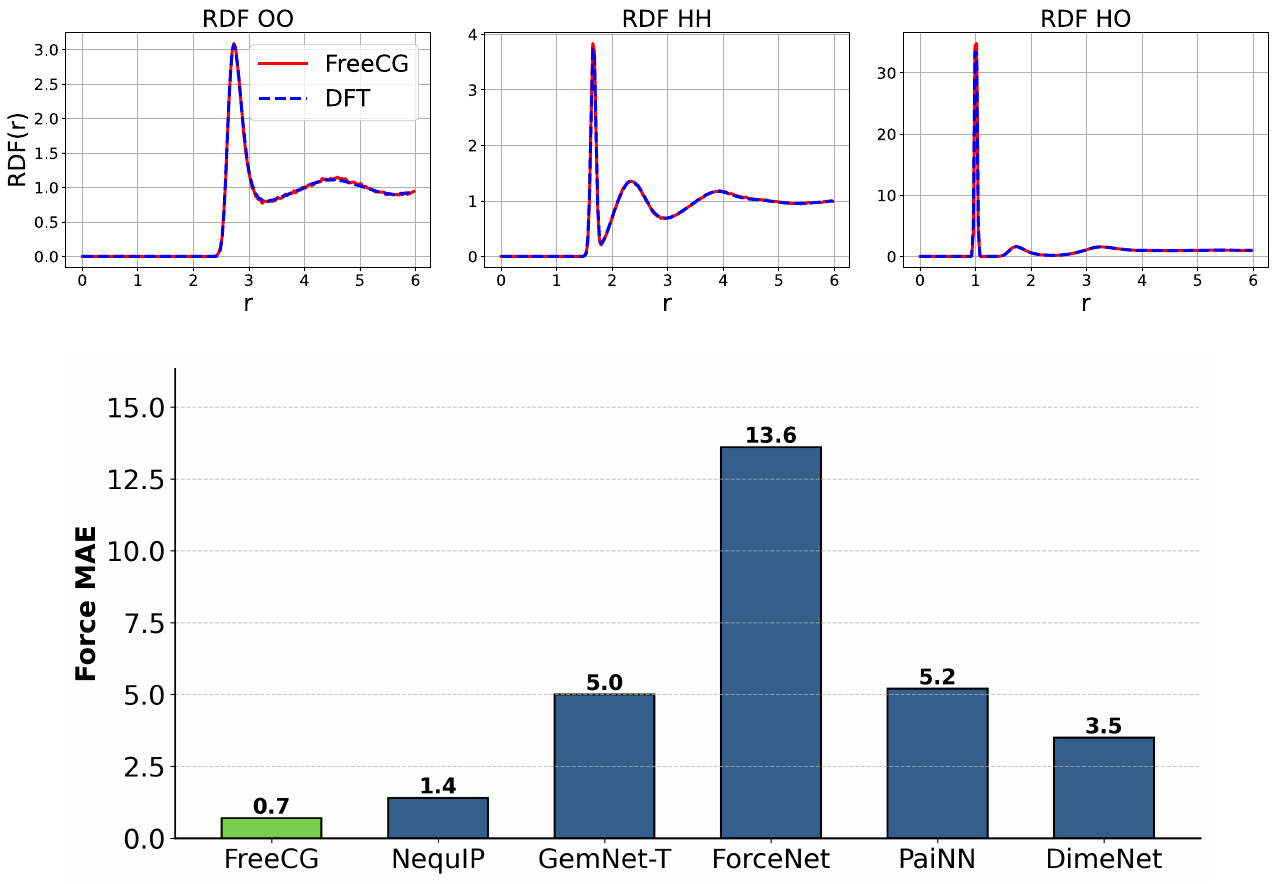}
  
  \caption{Results on water-1k. \textbf{Top:} The RDFs in molecular dynamic simulations for each bond of water under PBCs. The unit of $r$ is $\si{\angstrom}$, and the unit of ${\rm RDF}(r)$ is $\si{\angstrom}^{-1}$. \textbf{Bottom:} The force MAE comparison with other methods. The unit is meV/$\si{\angstrom}$.
  }
  \label{fig:mdwater}
\vspace{-0.2cm}
\end{figure*}

\begin{figure*}[tb]

  \centering
  \includegraphics[width=1.\linewidth]{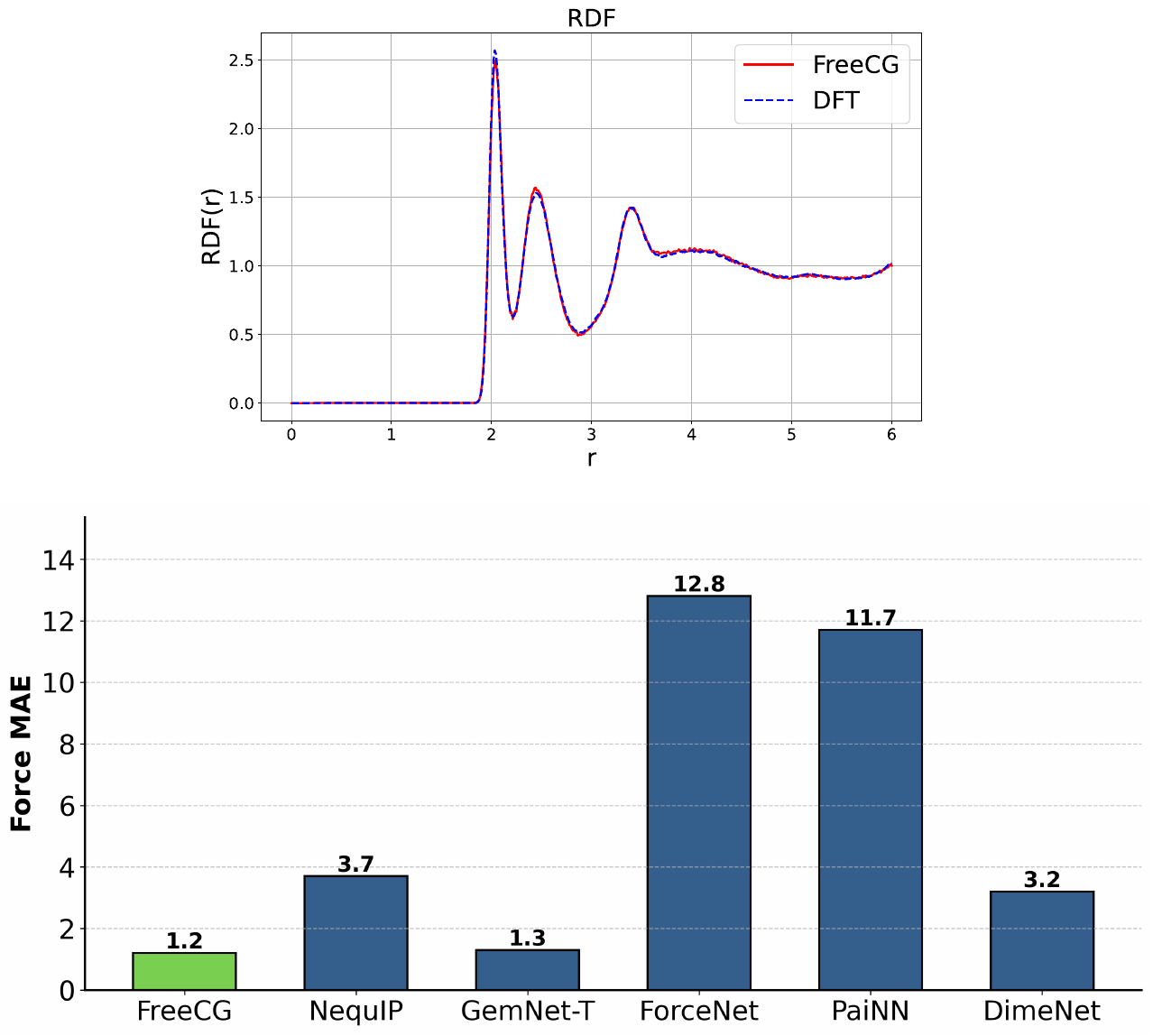}
  
  \caption{Results on LiPS. \textbf{Top:} The RDFs in molecular dynamic simulations for LiPS under PBCs. The unit of $r$ is $\si{\angstrom}$, and the unit of ${\rm RDF}(r)$ is $\si{\angstrom}^{-1}$. \textbf{Bottom:} The force MAE comparison with other methods. The unit is meV/$\si{\angstrom}$.
  }
  \label{fig:mdlips}
\vspace{-0.2cm}
\end{figure*}

\begin{figure*}[tb]
  
  \centering
  \includegraphics[width=1.\linewidth]{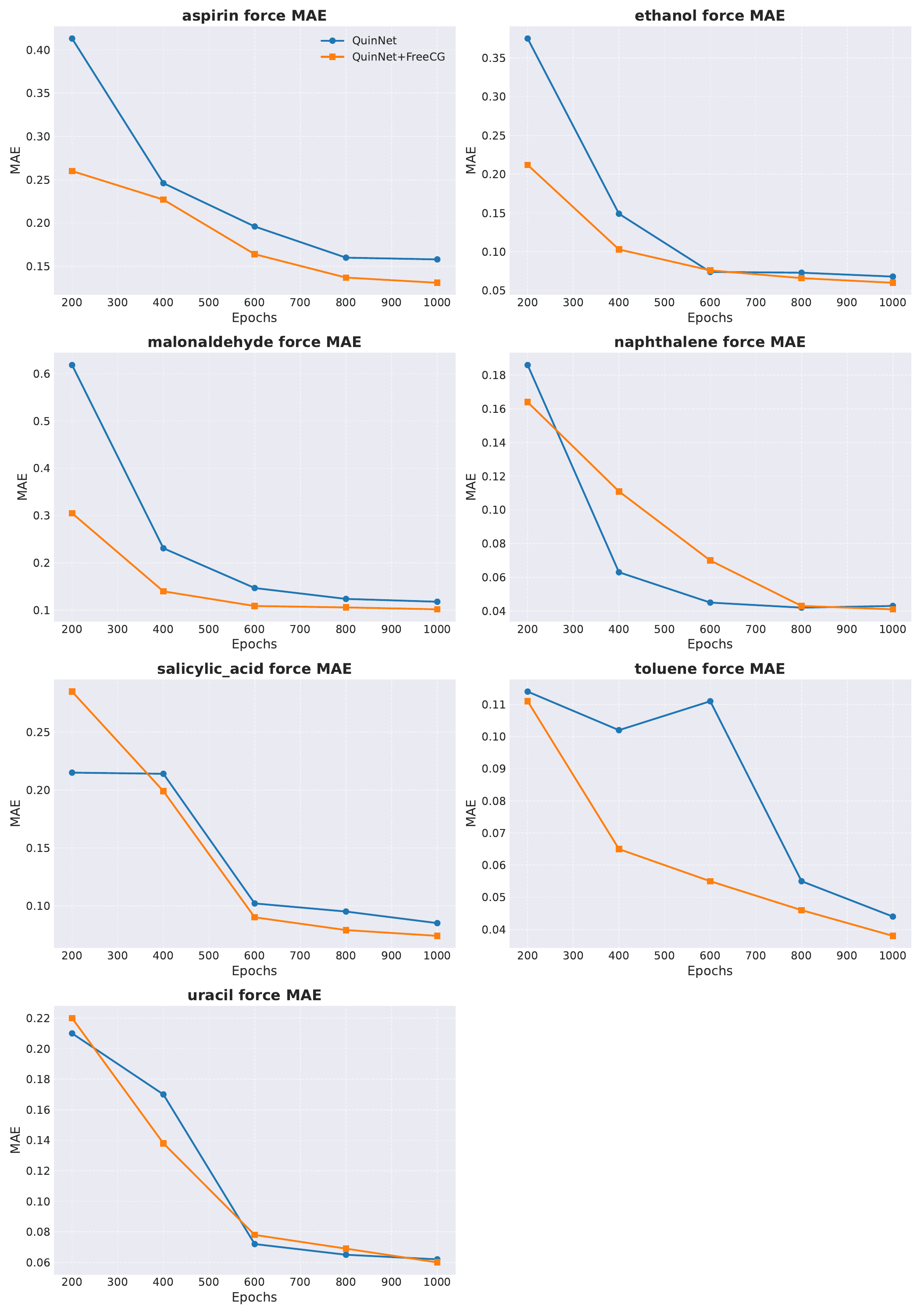}
  
  \caption{QuinNet force prediction performance on MD17 when equipped with modules from FreeCG. The unit of force is kcal/mol/$\angstrom$.
  }
  \label{fig:quinnetforce}
\vspace{-0.2cm}
\end{figure*}

\begin{figure*}[tb]
  
  \centering
  \includegraphics[width=1.\linewidth]{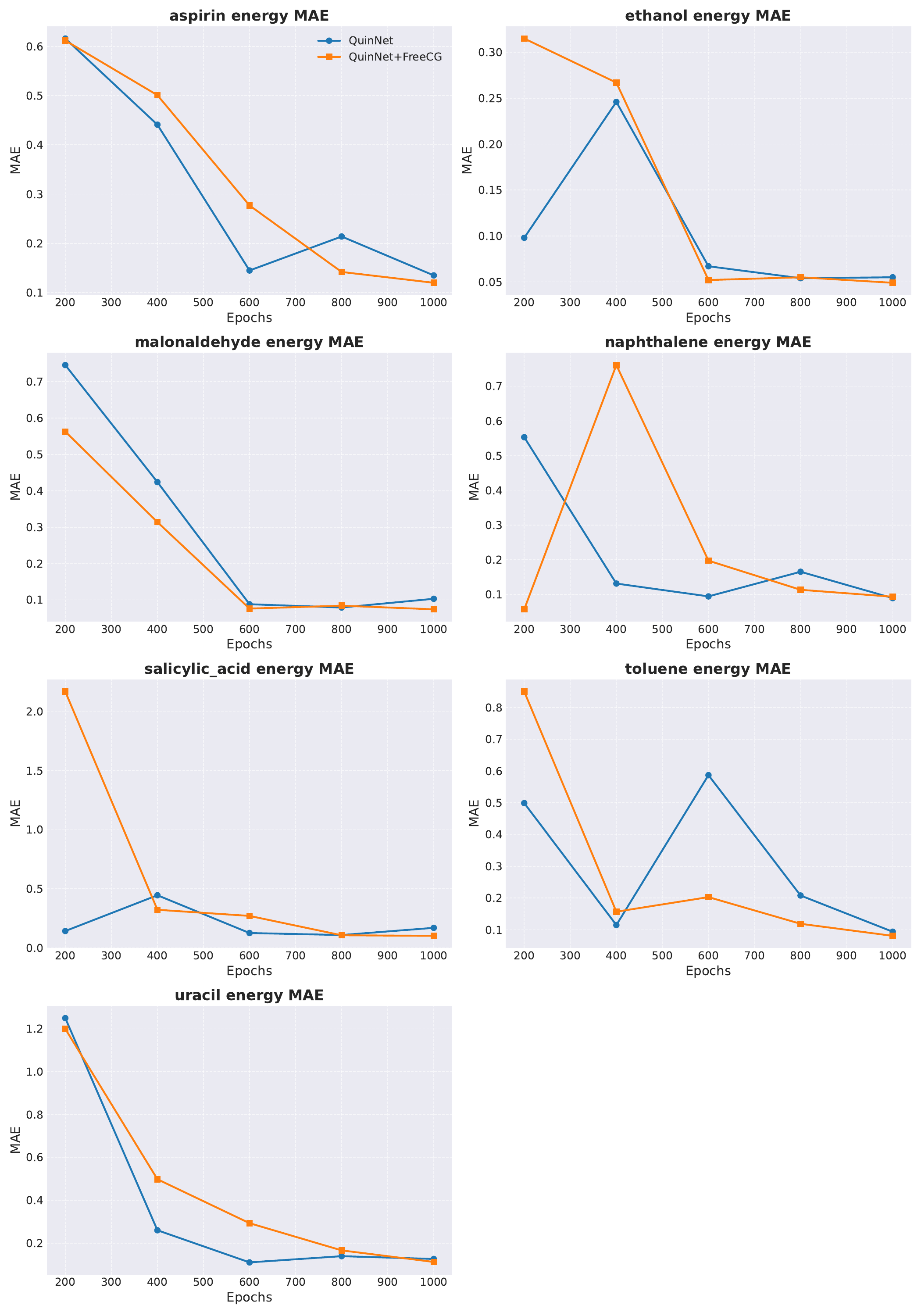}
  
  \caption{QuinNet energy prediction performance on MD17 when equipped with modules from FreeCG. The unit of energy is kcal/mol.
  }
  \label{fig:quinnetenergy}
\vspace{-0.2cm}
\end{figure*}

\end{document}